\documentclass[lettersize,journal]{IEEEtran}
\usepackage{amsmath,amsfonts}
\usepackage{algorithmic}
\usepackage{array}
\usepackage[caption=false,font=normalsize,labelfont=sf,textfont=sf]{subfig}
\usepackage{textcomp}
\usepackage{stfloats}
\usepackage{url}
\usepackage{verbatim}
\usepackage{graphicx}
\usepackage{footnote}
\def\BibTeX{{\rm B\kern-.05em{\sc i\kern-.025em b}\kern-.08em
    T\kern-.1667em\lower.7ex\hbox{E}\kern-.125emX}}
\usepackage{balance}
\usepackage{cite}
\usepackage{tabularx}

\begin{document}
\title{Radar Odometry for Autonomous Ground Vehicles: A Survey of Methods and Datasets}
\author{Nader J. Abu-Alrub, Nathir A. Rawashdeh,~\IEEEmembership{Senior Member,~IEEE}
\thanks{Nader J. Abu-Alrub is with the Department of Applied Computing, Michigan Technological University.}
\thanks{Nathir A. Rawashdeh is with the Department of Applied Computing, Michigan Technological University.}
}

\maketitle

\begin{abstract}
Radar odometry has been gaining attention in the last decade. It stands as one of the best solutions for robotic state estimation in unfavorable conditions; conditions where other interoceptive and exteroceptive sensors may fall short. Radars are widely adopted, resilient to weather and illumination, and provide Doppler information which make them very attractive for such tasks. This article presents an extensive survey of the latest work on ground-based radar odometry for autonomous robots. It covers technologies, datasets, metrics, and approaches that have been developed in the last decade in addition to in-depth analysis and categorization of the various methods and techniques applied to tackle this problem. This article concludes with challenges and future recommendations to advance the field of radar odometry making it a great starting point for newcomers and a valuable reference for experienced researchers.
\end{abstract}


\section{Introduction}
\label{sec:introduction}
\IEEEPARstart{R}{adars} have been an essential component of the perception systems in many autonomous platforms. They are proven technology and have been widely used in many related robotic applications. Range measurement, resilience, and Doppler information are examples of the qualities giving them an important role in these applications. 

Odometry is usually defined as the process of estimating the ego-motion of a mobile platform with respect to a fixed reference point using sensors such as wheel encoders, Inertial Measurement Unit (IMU), camera, lidar, or radar. This ego-motion is typically described as pose information (i.e., position and orientation) and/or velocity. Odometry is crucial in environments where the Global Navigation Satellite System (GNSS) signal is blocked or not reliable enough to estimate the location of a moving vehicle. In common robotic tasks such as obstacle avoidance, object tracking, path planning, and navigation, a mobile robot is expected to keep track of its position and orientation with respect to a fixed frame in order to track and localize other objects in its vicinity. This is typically done with the help of the GNSS. For applications where GNSS is not an option, odometry stands out as the best alternative. Examples of such scenarios are Urban and off-road driving, mining and subterranean robots, underwater and subsea applications, and space exploration. Odometry is different from localization where a priori knowledge in the form of a map is used to describe the pose of a robot. It is also different from Simultaneous Localization and Mapping (SLAM) where a robot is actively building a map and localizing in that map, offline or online, while odometry algorithms are not meant to build maps and are expected to always run in realtime.

\begin{figure}[!t]
    \centering
    \includegraphics[width=3.2in]{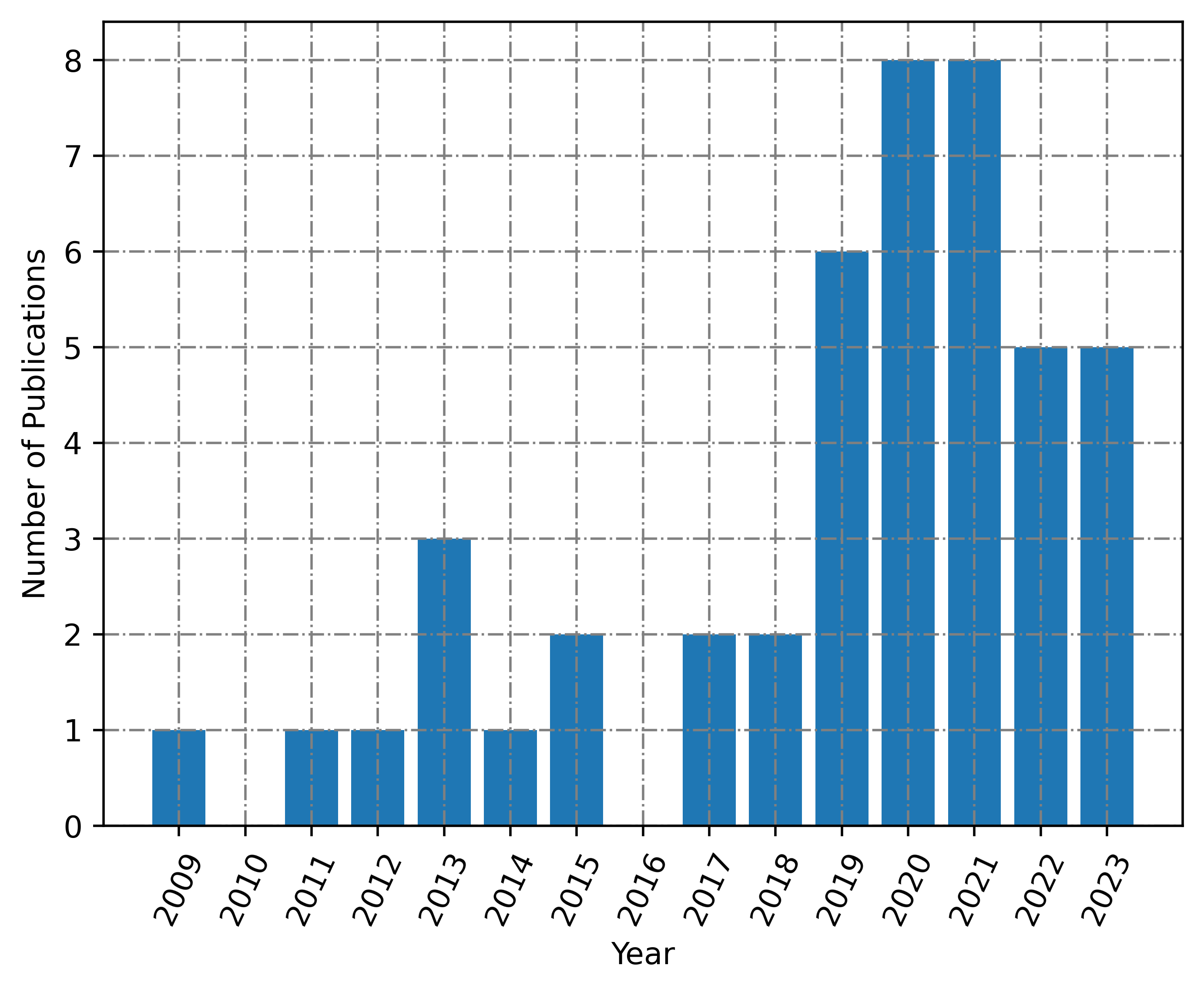}
    \caption{Relevant publications to radar odometry for autonomous ground vehicles in the last decade. The last bar is as of June, 2023.}
    \label{fig:publication_count}
\end{figure}

Radar odometry refers to the use of radar sensors as the primary source of measurements for estimating the ego-motion of a mobile platform. The use of radar for odometry has been gaining traction in the last decade (see Fig. \ref{fig:publication_count}) for many reasons; radars' resilience to challenging weather conditions compared to cameras and lidars is one of the most notable ones. Thanks to its longer operating wavelengths, radars are far less affected by rain, dust, fog, and snow \cite{zhang2023weather1, mohammed2020weather2, bijelic2018weather3}. Radars are also not affected by lighting and illumination source like cameras which give them an edge in scenarios such as night, cloudy days, underwater, and mineshafts. Other advantages of radar include its suitability for indoor and outdoor applications (unlike GNSS-based localization), it does not suffer from slip or skid problems as wheel encoders do, and it is less susceptible to drift as IMUs are. Moreover, radars have long range, may provide velocity measurements, are relatively cheap, compact, and already widely adopted in the autonomous vehicles market.

There have been few attempts to compile radar odometry related work, the published surveys \cite{survey_automotive_radar_key,survey_radar_automotives,survey_odometry_autonomous,survey_sensors_indoor_odometry} each provide a section on radar odometry, its categories, sensor types, and latest methods; however, they are not dedicated to radar odometry and therefore, miss a lot of relevant publications.  The work in\cite{survey_radar_uav} has a section on radar odometry but it only treats Unmanned Aerial Vehicles (UAV) cases. Additionally, \cite{survey_radar_localization} gives attention to localization and SLAM and forgoes the odometry steps. The only recent survey on radar odometry is the work by Lubanco \textit{et al.} \cite{Lubanco2022radarsurvey} which offers a very brief overview of the topic and falls short in terms of covering and analyzing all the recent literature on radar odometry. In this survey, we attempt to cover this gap and compile all recent work that is related to ground-based radar odometry in one extensive article for researchers and newcomers to quickly and easily get acquainted with the most essential knowledge needed to study and contribute to the field. We analyze published research on the topic of radar odometry in the last decade or so, few exceptions were made for seminal work dated before that. Generally, we are looking for purely odometry pipelines, but we have included some localization (pose estimation with a priori knowledge) and SLAM work where the odometry pipeline, to the best of our judgment, is distinguishable from other stages in the process. We narrow down our scope to Unmanned Ground Vehicles (UGV) and exclude UAVs. We believe that the more stringent runtime and payload requirements and the more relaxed spatial accuracy associated with UAV odometry make it a quite different problem with a unique set of strategies and solutions worthy of its dedicated survey. Finally, we discuss fusion-based approaches that rely on radars and a mix of other sensors in a separate section, section \ref{sec:fusion-based_radar_odometry}, even though these methods might not be considered pure radar odometry implementations. 

The remainder of this paper is organized as follows: section \ref{sec:radar_sensor} provides an overview of the principles of radar sensors used in robotic applications. We present a general formulation of the odometry problem in section \ref{sec:odometry_formulation} followed by the evaluation metrics commonly used to assess radar odometry in section \ref{sec:metrics}. Section \ref{sec:radar_dataset} is dedicated to datasets featuring radar data that can be used for radar odometry research. Section \ref{sec:radar_odometry} is an extensive survey of the literature on radar odometry and its different categories. Sections \ref{sec:fusion-based_radar_odometry} and \ref{sec:radar_odometry_and_deep_learning} discuss the use of different fusion and machine learning techniques in radar odometry, respectively. Finally, we highlight some challenges and future recommendations relevant to radar odometry in section \ref{sec:problems&recommendaitons} and conclude this article in section \ref{sec:conclusion}.

\section{Princibles of mmWave Radars}
\label{sec:radar_sensor}
\noindent
mmWave radars are characterized by utilizing signal frequencies in the range of 30-300 GHz which translates to wavelengths in the range of 1-10 mm; however, the frequency range commonly used for automotive radars is the narrower 70-100 GHz range. The advantages of using this range of wavelengths are the reduced cost and size of hardware and improved accuracy \cite{ti_mmwave}. Radars are already widely adopted in autonomous vehicles and robotics applications such as blind spot detection, rear collision warning, adaptive cruise control, object detection, tracking, and classification, in addition to odometry, localization, and mapping. This section briefly explains the basic principles of mmWave Frequency Modulated Continuous Wave (FMCW) that are typically used in automotive and robotic systems, readers interested in further details are referred to \cite{adams2012book} and \cite{gamba2019book} for more thorough analysis on mmWave radars in automotive/robotic applications.

\begin{figure*}[!t]
    \centering
    \subfloat[]{\includegraphics[width=1.5in]{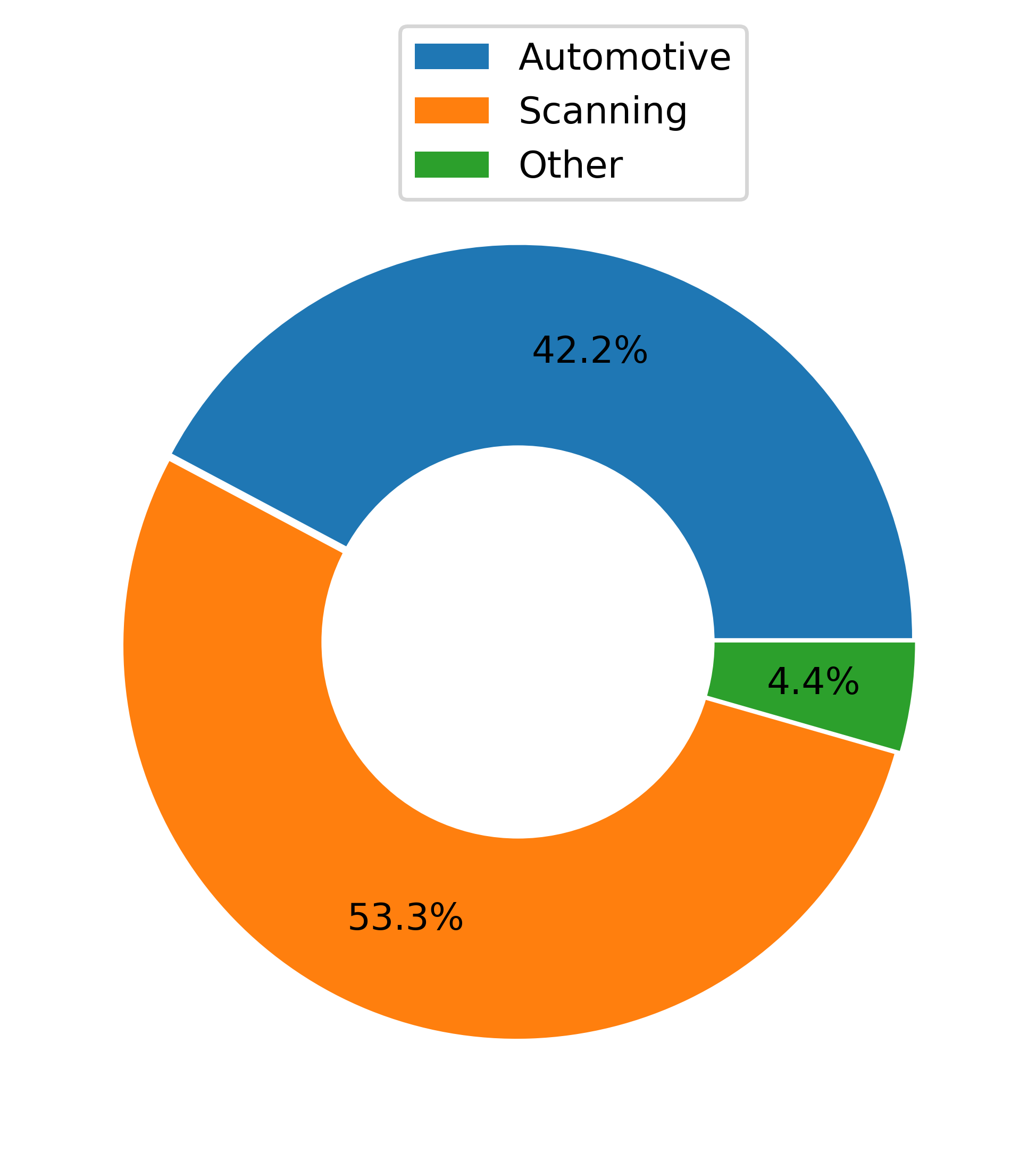}
    \label{fig:automotive_vs_scanning}}
    \hfil
    \subfloat[]{\includegraphics[width=1.5in]{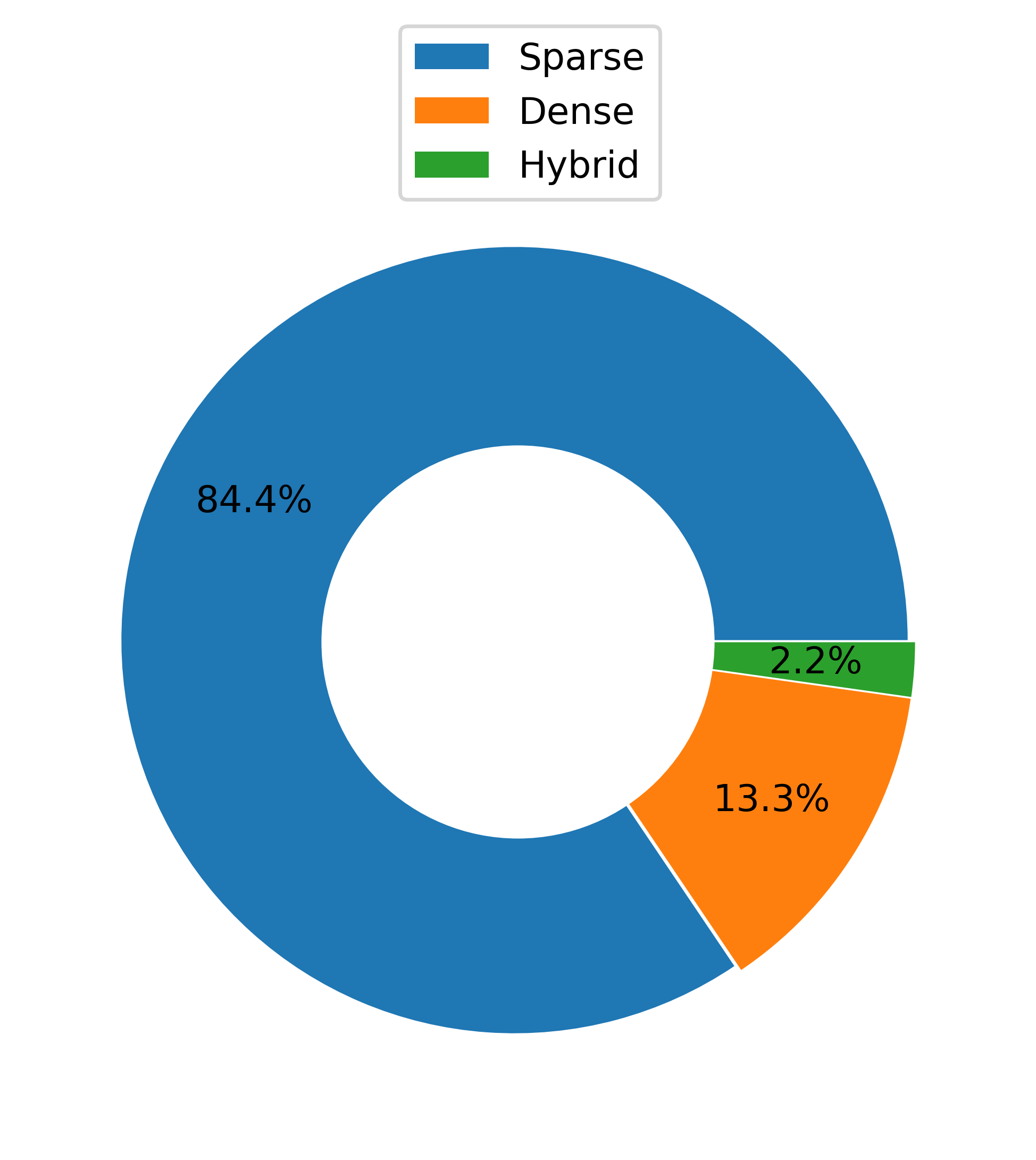}
    \label{fig:sparse_vs_dense}}
    \hfil
    \subfloat[]{\includegraphics[width=1.5in]{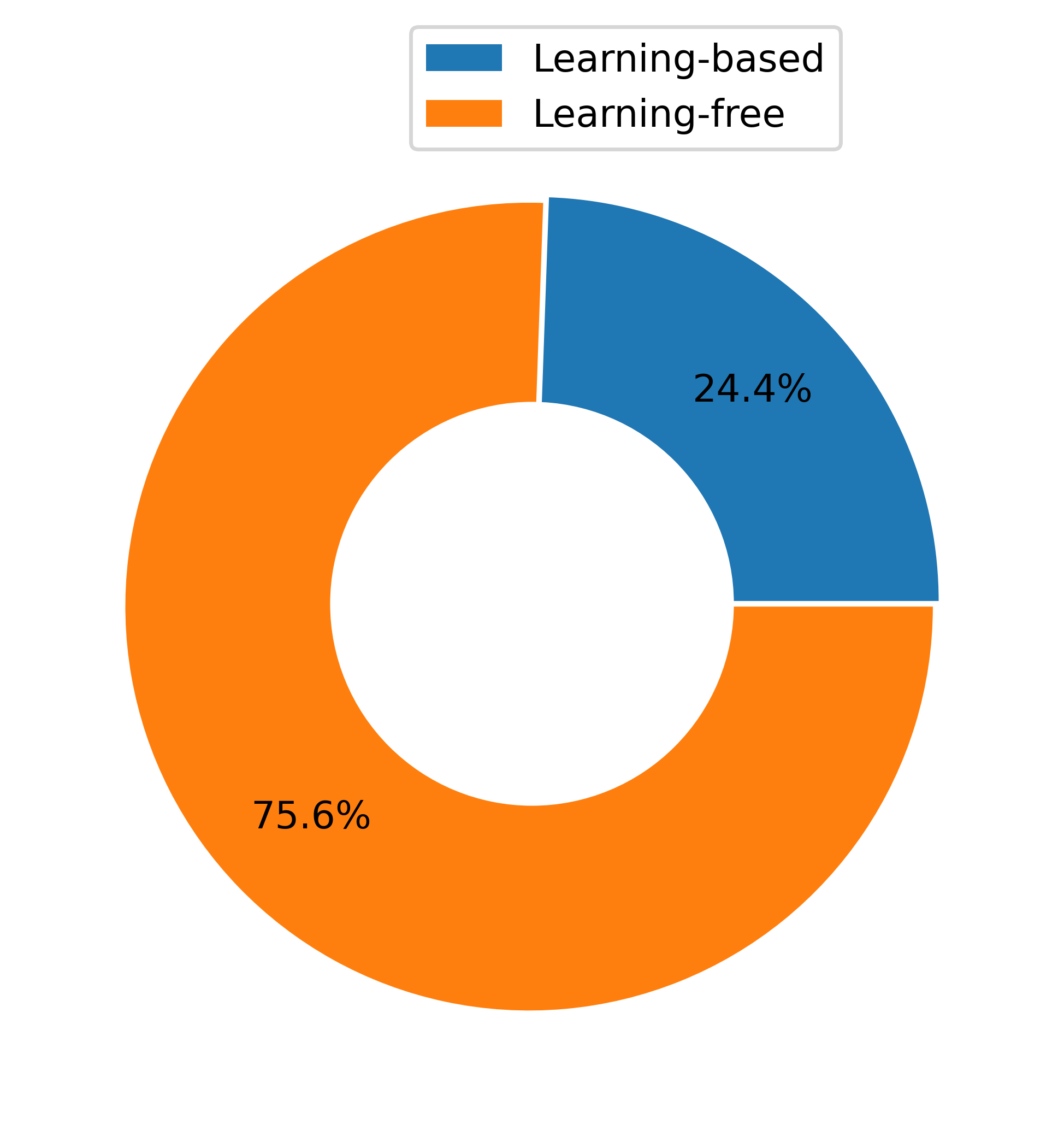}
    \label{fig:learning_vs_nolearning}}
    \hfil
    \subfloat[]{\includegraphics[width=1.5in]{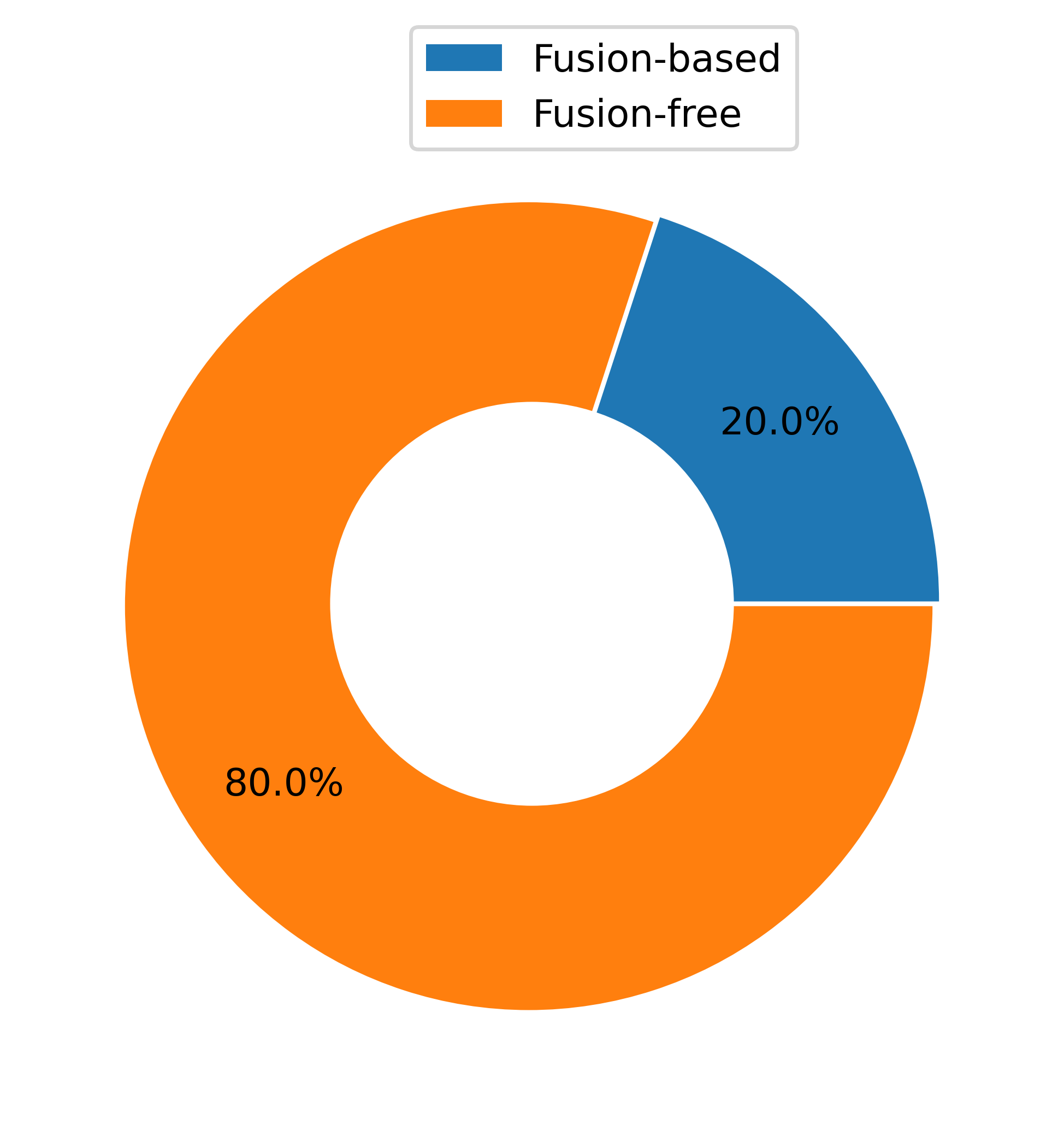}
    \label{fig:fusion_vs_nofusion}}
    \hfil
    \caption{Trends in radar odometry for autonomous ground vehicles in the last decade. (a) Usage of automotive, scanning, and other types of radars. (b) The popularity of sparse, dense, and hybrid methods. (c) Learning-based vs Learning-free methods in radar odometry. (d) Fusion-based vs Fusion-free methods in radar odometry.}
    \label{fig:survey_stats_pie_charts}
\end{figure*}

\begin{figure}[!t]
    \centering
    \subfloat[]{\includegraphics[width=1.6in]{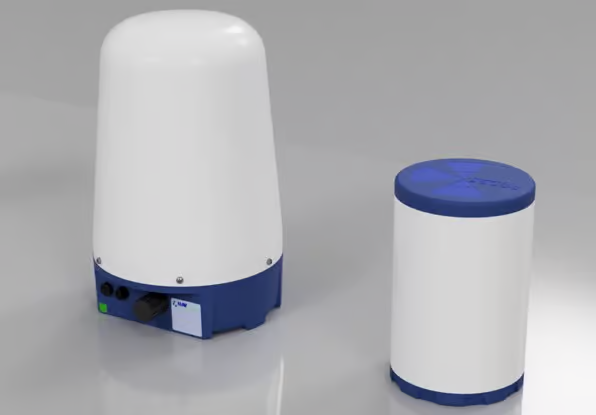}
    \label{fig:navtech_radar_example}}
    \hfil
    \subfloat[]{\includegraphics[width=1.6in]{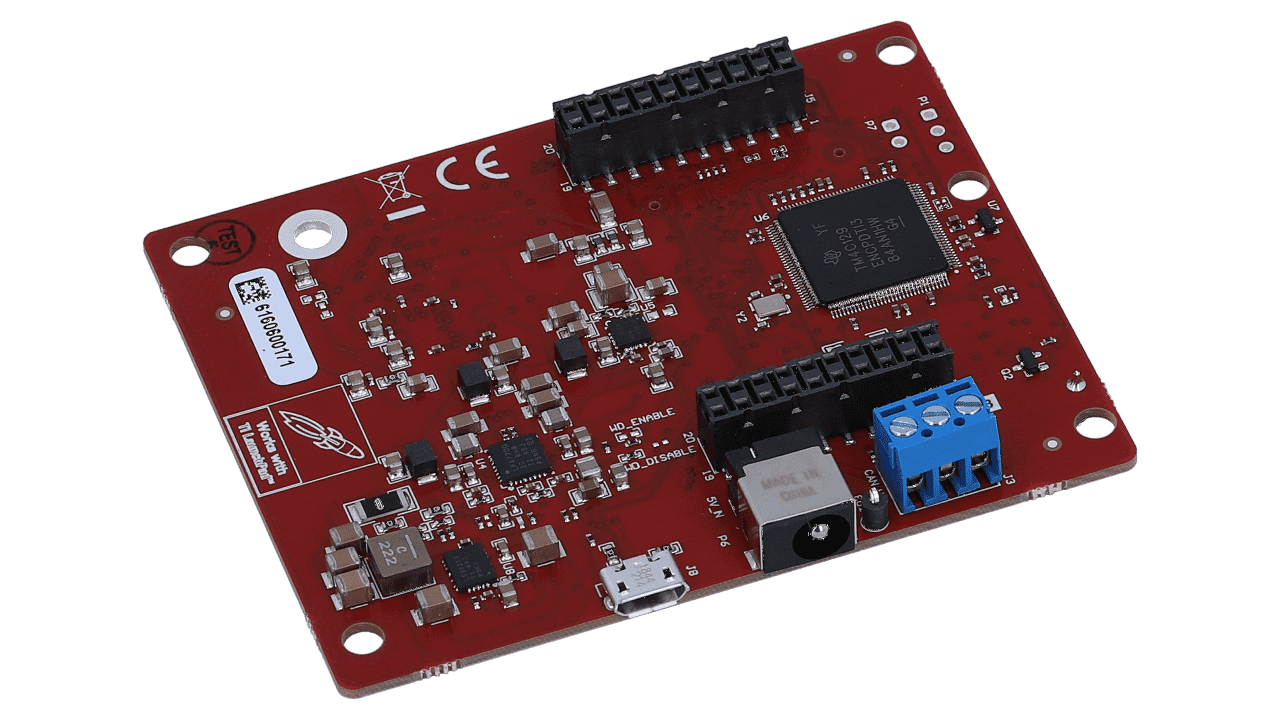}
    \label{fig:TI_radar_example}}
    \caption{Examples of radar sensors used for odometry and autonomy applications. (a) Navtech CIR and CAS scanning radar sensors. Source:\cite{navtech_website}. (b) Texas Instruments AWR1843BOOST automotive radar sensor. Source:\cite{TI_website}.}
    \label{fig:radar_examples}
\end{figure}

The literature on radar odometry typically distinguishes between two broad types of radars, the so-called "automotive radar", which generates a sparse point cloud of detections usually returned as a list of range, radial velocity, azimuth, and elevation per target detection. The other type is the "scanning radar", which returns dense $360^o$ Bird Eye View (BEV) scans. Fig. \ref{fig:automotive_vs_scanning} shows the percentages of published work based on each type of radars. There are multiple examples of radar odometry approaches based on custom-built scanning radars in the literature (e.g., \cite{fritsche2017icp2,marck2013icp1}); however, the vast majority of research utilizing scanning radars is based on the radars developed by Navtech \cite{navtech_website}, therefore, extra attention will be given to these. Fig. \ref{fig:radar_examples} shows an example of both types of radars.

Commonly used in automotive and robotics applications are the FMCW radars. Unlike pulsed radars, FMCW radars transmit and receive a predetermined range of frequencies continuously, whereas, in pulsed radars, high-energy pulses are transmitted/received. The time required for a pulse to complete a round trip (i.e., Time of Flight (ToF)) is measured and used to calculate the distance to the object that reflected the pulse. One way to calculate radial velocity in pulsed radars is by recording two consecutive timed range measurements and calculating the rate of change of the distance. Obtaining these measurements in FMCW radars is not as straightforward; FMCW radars work by transmitting two continuous orthogonal sinusoidal signals that are 90 degrees shifted, called In-phase and Quadrature (IQ) signals as in equations (\ref{equ:signal_a}) and (\ref{equ:signal_b}):

\begin{equation}
    S_{I,tx} = sin(f_{tx}t)
    \label{equ:signal_a}
\end{equation}
\begin{equation}
    S_{Q,tx} = sin(f_{tx}t-\frac{\pi}{2})
    \label{equ:signal_b}
\end{equation}

Where $f_{tx}$ is the transmit frequency. $S_I$ and $S_Q$ are the In-phase and Quadrature signals, respectively. When reflected from an object, depending on the relative motion of the object with respect to the radar, the frequency of the signal will be shifted due to the Doppler effect. When received by the radar, the radar mixes the transmitted/received In-phase signals and the transmitted/received Quadrature signals by multiplying them. For example, for the In-phase signals:

\begin{equation}
    S_{I,tx} \cdot S_{I,rx} = sin(f_{tx}t) \cdot sin(f_{rx}t)
    \label{equ:signal_mixing}
\end{equation}

Simplifing equation (\ref{equ:signal_mixing}) gives the \textit{beat frequency} ($f_{tx}-f_{rx}$) as in equation (\ref{equ:beat_freq}) where $f_{rx}$ is the received frequency. 

\begin{equation}
    S_{I,tx} \cdot S_{I,rx} = \frac{1}{2}[cos((f_{tx}-f_{rx})t) -cos((f_{tx}+f_{rx})t)]
    \label{equ:beat_freq}
\end{equation}

This \textit{beat frequency} can be used to calculate the radial velocity of an object. The direction of the velocity can be determined by checking which of the IQ signals is leading/lagging.

For distance measurement, modulation of the frequency is required. In its most basic form, Linear Frequency Modulated Continuous Wave (LFMCW) ramps the transmitted signal to produce a chirp. This chirp, once received back with a delay, produces a beat frequency that can be used to estimate the distance. Problems arise when both velocity and range estimates are required which can be solved using other frequency modulations such as sawtooth, which is, again, limited when there are multiple objects in the scene. This problem is typically addressed using more sophisticated techniques such as Multiple Frequency Shift Keying (MFSK).
Finally, another measurement of interest for autonomous driving is the angle of the detected object, named the Angle of Arrival (AoA), using multiple receive antennas displaced by a known distance, we can exploit the phase shift in the received signals among antennas to estimate the angle of the object, of course, the angular resolution determines how close objects can be located next to each other and still considered separate. The angular resolution can be improved by adding more receive antennas and even more transmitters to get Multi-Input Multi-Output (MIMO) radars which can increase the accuracy of radar systems without increasing the size of the array beyond the reasonable size.

\begin{figure}[!t]
\centering
\subfloat[]{\includegraphics[width=1.6in]{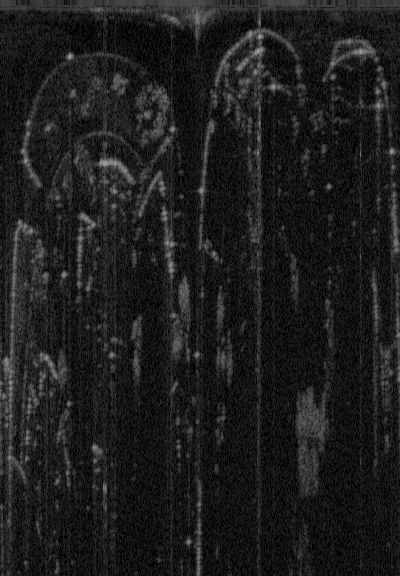}
\label{fig:polar2cartesian_a}}
\hfil
\subfloat[]{\includegraphics[width=1.6in]{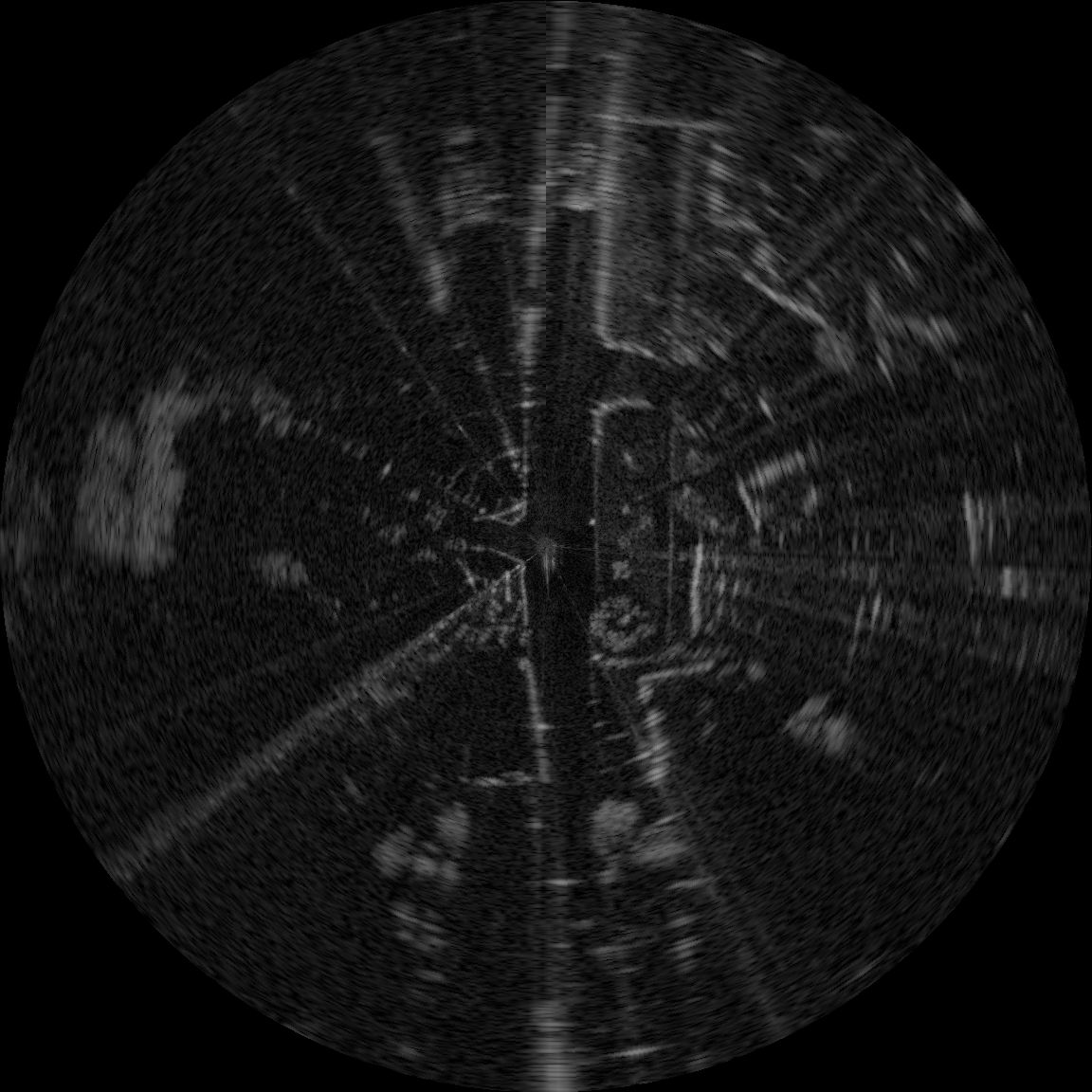}
\label{fig:polar2cartesian_b}}
\caption{An example of a single radar scan in polar and cartesian coordinates representation taken from \textit{RADIATE} dataset \cite{radiate_dataset}. (a) Polar coordinates. (b) Cartesian coordinates.}
\label{fig:polar2cartesian}
\end{figure}

Typical scanning radars are made to rotate around a vertical axis while continuously transmitting and receiving signals. The power of the reflected signals depends on the reflectivity of these objects, their size, and their range. At each azimuth angle $N$, a 1D signal is received with power readings corresponding to $M$ number of range bins. These readings form an image in polar coordinates where points can be represented as $p(a,r)$ where $a$ and $r$ denote azimuth and range of point $p$. It is usually desirable to map and process these radar images into cartesian coordinates in which the vehicle (or the radar) appears in the middle of the image. This can be done by projecting all points from polar coordinates to cartesian coordinates using a mapping function similar to (\ref{equ:coordinates_mapping}):

\begin{equation}
    \label{equ:coordinates_mapping}
{p = }
\begin{bmatrix}
    \mu \cdot r \cdot cos\theta  \\ \mu \cdot r \cdot sin\theta 
\end{bmatrix}
\end{equation}

where $\theta = a . \frac{2\pi}{N}$ and $\mu$ is a scale factor (m/pixel). Gaps in the cartesian image are usually filled using interpolation. An example of a radar scan in polar and cartesian coordinates is shown in Fig. \ref{fig:polar2cartesian}.

\begin{figure}[!t]
\centering
\includegraphics[width=2.5in]{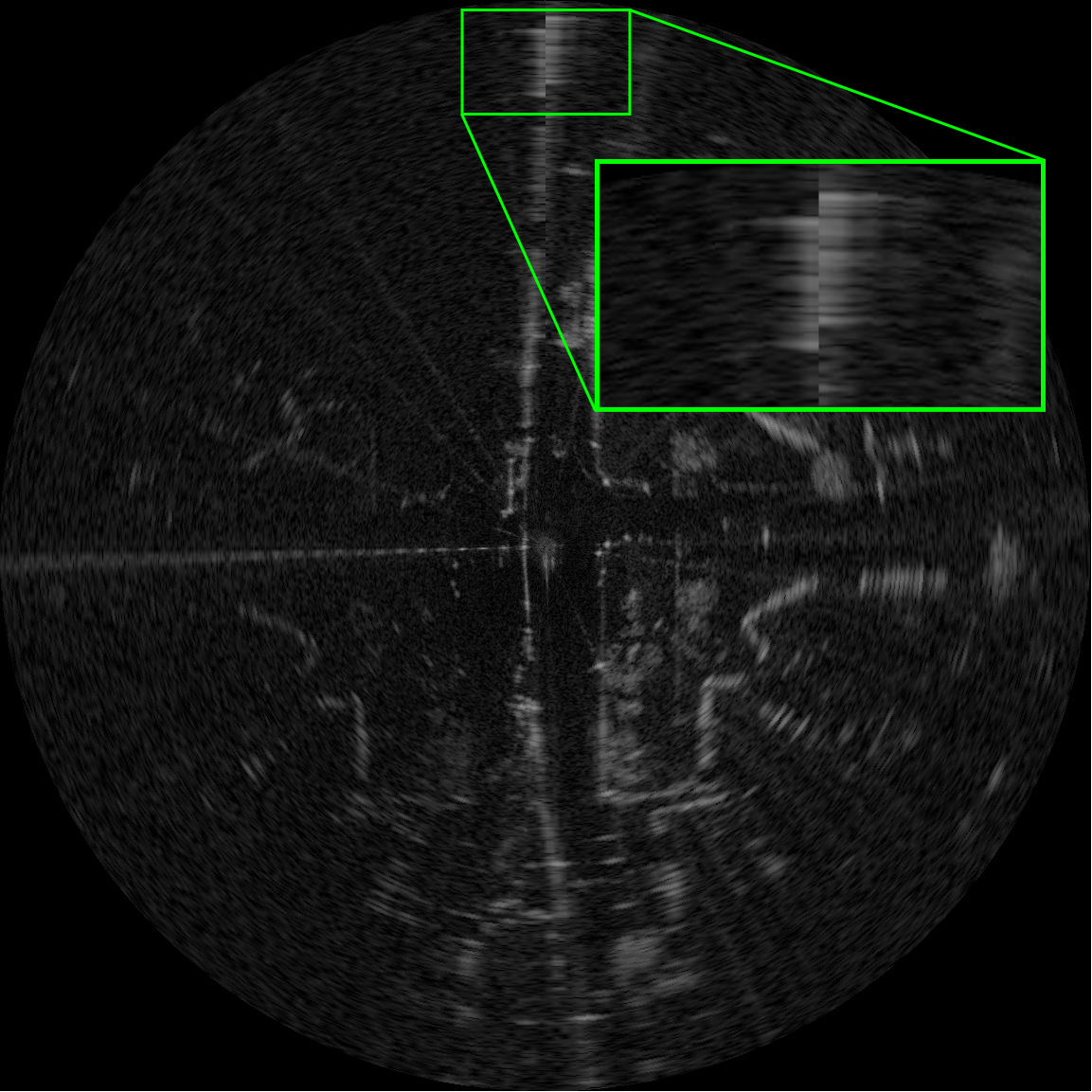}
\caption{An example of a distorted radar scan taken from \textit{RADIATE} dataset \cite{radiate_dataset}. A detection in front of the radar, probably another vehicle, appears broken as a result of being detected at the beginning of the scan and the end of the same scan (i.e., approximately 0.25 seconds later).}
\label{fig:distortion}
\end{figure}

A common problem that arises in scanning radars and other "recording-while-moving" types of sensors is the motion distortion in the resulting scan. Although this problem exists in other sensors, for example, in rolling shutter cameras \cite{rolling_shutter} and scanning lidars \cite{KISS-ICP}, it is more prevalent in scanning radar sensors because of their relatively slow scan rate. A scanning radar like the models from Navtech has a sampling frequency of 4Hz, a full scan would take 0.25 seconds to complete. Assuming that the sensor is mounted on a car moving at a moderate speed of 30 mph (or 48.3 km/h), by the time a full scan is completed, an object in front would have moved 11 ft (or 3.35 m) closer, which is a car-sized difference in the range estimate, something that can be seriously dangerous. Burnett \textit{et al.} \cite{motion_compensation} provide great details on the effect of motion distortion on odometry and localization using scanning radar and highlight the importance of compensating for these artifacts. Fig. \ref{fig:distortion} shows an example of a distorted radar scan. The most common solution for this problem is to perform motion distortion compensation (or scan deskewing) which relies on estimating the shift in the location of points and offset for it based on the assumption that the velocity is constant while taking a single scan.

Despite the added complexity when compared to pulsed radars, FMCW radars are still favored in autonomous vehicles and most robotic applications because of their smaller form factor, lower power consumption, and lower cost, in addition to the fact that they are not restricted by a relatively far minimum range of detection.

\section{General Problem Formulation}
\label{sec:odometry_formulation}
\noindent

This section presents a general formulation of the odometry problem in its most basic form. This formulation is not unique to radar odometry and can be applied to algorithms based on other sensors (e.g., visual and laser). For the sake of generality, the problem will be described here in $SE(3)$ instead of $SE(2)$ even though the 2D version is usually sufficient for applications in autonomous vehicles and many ground-based robots where the z-dimension, roll, and pitch are usually ignored.

Assuming a mobile robot is moving in its environment and collecting radar data at discrete time instants $k$ where $k=\{0, 1, 2, ... n\}$ and $n$ is the total number of scans at the end of the trajectory. The robot has an initial pose $P_0$ which can be set arbitrarily by the user or aligned with an external source. At each time instant $k$, a sensor scan, denoted by $S_k$, is recorded, the set of all sensor scans throughout the journey is $S_{k:n} = \{S_0, S_1, S_2, ... S_n\}$. The robot's poses can be expressed with respect to the initial pose by $^0P_k$. The set of all robot poses throughout the journey is then $^0P_{k:n} = \{^0P_0, ^0P_1, ^0P_2, ... ^0P_n\}$. One way to express these poses in 3D using homogeneous coordinates is as in  (\ref{equ:pose_def}) combining orientation (as rotation matrix $\textbf{R}^{3\times3}_k$) and position (as translation vector $\textbf{t}^{3\times1}_k$):
\begin{equation}
    \label{equ:pose_def}
    {P_k = }
    \begin{bmatrix}
        \textbf{R}^{3\times3}_k & \textbf{t}^{3\times1}_k  \\ \textbf{0}^{1\times3} & 1 
    \end{bmatrix}
\end{equation}
Two consecutive poses, $P_{k-1}$ and $P_k$, are related by the rigid body transformation:
\begin{equation}
    \label{equ:rigid_body_transformation}
    {^{k-1}T_k = }
    \begin{bmatrix}
        ^{k-1}\textbf{R}_k & ^{k-1}\textbf{t}_k  \\ \textbf{0} & 1 
    \end{bmatrix}
\end{equation}
where $^{k-1}T_k \in \mathbb{R}^{4x4}$, $^{k-1}\textbf{R}_k \in SO(3)$ is the rotation matrix, and $^{k-1}\textbf{t}_k \in \mathbb{R}^{3x1}$ is the translation vector. The main goal of odometry is to calculate the relative transformation, $^{k-1}T_k$,  between adjacent pairs of poses in order to retrieve the set of all transformations which describes the full motion throughout the journey:
\begin{equation}
    \label{equ:transformations_set}
    ^{k:n-1}T_{k+1:n} = \{^0T_1, ^1T_2, ^2T_3, ... ^{n-1}T_n\}
\end{equation}
The current pose of the robot with respect to the starting point $^0P_k$ can be continuously updated by incrementally concatenating relative transformations using (\ref{equ:poses_concatentaion}):
\begin{equation}
    \label{equ:poses_concatentaion}
    ^0P_k =   ^0P_{k-1} \ ^{k-1}T_k
\end{equation}
Finally, the set of all absolute poses that fully describes the trajectory is: 
\begin{equation}
    \label{equ:poses_set}
    ^0P_{k:n} = \{^0P_0, ^0P_1, ^0P_2, ... ^0P_n\}
\end{equation}

It is worth mentioning here that the terms "odometry" and "ego-motion estimation" are sometimes used interchangeably in the literature. The most common distinction between them is that "ego-motion estimation" typically refers to solving for the relative transformation between two poses described in (\ref{equ:rigid_body_transformation}) whereas the term "odometry" usually refers to the recovery of the complete trajectory up to the current instance in time as described in (\ref{equ:poses_set}).

\section{Odometry Evaluation Metrics}
\label{sec:metrics}
\noindent
Crucial for benchmarking odometry algorithms is to understand their evaluation metrics. This section presents a brief explanation of the most popular odometry metrics in the literature. These metrics are general and not specific for radar-based approaches, i.e., they can be used to evaluate odometry algorithms based on visual, laser, or any other sensor. 

\subsection{Absolute Trajectory Error (ATE)}
\label{subseq:ate}
\noindent
The Absolute Trajectory Error (ATE) is the sum of the root mean square errors along a trajectory as in equation (\ref{equ:ate}). The error at each pose is calculated using the Euclidean distance between the estimated position $t_{k,est}$ and the ground truth position $t_{k,gt}$ using equation (\ref{equ:pose_error}).

\begin{equation}
    \label{equ:ate}
    ATE = \sqrt{\frac{1}{n}\sum_{k=0}^n{e_k}^2}
\end{equation}

\begin{equation}
    \label{equ:pose_error}
    e_k = \sqrt{\sum{({t_{k,est}} - {t_{k,gt}})^2}}
\end{equation}

The ATE metric is straightforward and provides a quick and easy assessment of the algorithm at hand. Limitations of the ATE include the fact that it does not hint to where or when did the wrong pose estimation take place or whether it was an error in the translational or the rotational component. Finally, ATE accumulates tiny errors in orientation estimates to huge ATE values as the journey progresses.

\subsection{Relative Pose Error (RPE)}
\label{subseq:rpe}
\noindent
Relative Pose Error has two separate components describing the relative error in translation and relative error in rotation. The idea is that instead of calculating the error between two corresponding poses $P_{k,est}$ and $P_{k,gt}$  at some discrete time instant $k$, the error is calculated between the transformations that align $P_{k-1,est}$ with $P_{k,est}$ and $P_{k-1,gt}$ with $P_{k,gt}$, respectively. These intermediate relative transformations are calculated using:

\begin{equation}
    ^{k-1}\delta_{k,est} = \space P_{k-1,est} \ominus \space  P_{k,est}
\end{equation}

\begin{equation}
    ^{k-1}\delta_{k,gt} = \space P_{k-1,gt} \ominus \space P_{k,gt}
\end{equation}

Where $\ominus$ is the inverse motion composition operator. For example, $\delta_{a,b} = x_b \ominus x_a$ is the relative transformation that aligns $x_a$ with $x_b$. The operator $\ominus$ is the inverse of the standard motion composition operator $\oplus$ \cite{kummerle2009metrics}. 

The translational RPE measures the difference (i.e., misalignment) between the translational components of $^{k-1}\delta_{k,est}$ and $^{k-1}\delta_{k,gt}$. Similarly,  the rotational RPE measures the deviation between the rotational components of $^{k-1}\delta_{k,est}$ and $^{k-1}\delta_{k,gt}$. The average of all these relative errors is calculated along the trajectory using (\ref{equ:trans_of_rpe}) and (\ref{equ:rot_of_rpe}):

\begin{equation}
    RPE_{trans} = \frac{1}{n}\sum_{k=1}^n {trans(^{k-1}\delta_{k,est} \ominus \space ^{k-1}\delta_{k,gt})}
    \label{equ:trans_of_rpe}
\end{equation}

\begin{equation}
    RPE_{rot} = \frac{1}{n}\sum_{k=1}^n {rot(^{k-1}\delta_{k,est} \ominus \space ^{k-1}\delta_{k,gt})}
    \label{equ:rot_of_rpe}
\end{equation}

Where the functions $trans(.)$ and $rot(.)$ are used to extract the translational and rotational components of a relative transformation matrix. For example, for an arbitrary relative transformation matrix $^{k-1}T_k$ (similar to (\ref{equ:rigid_body_transformation})), the translational error is calculated using the $L_2-norm$ of the translation vector $t$:

\begin{equation}
    trans(T) = ||t||
\end{equation}

and the rotational error is calculated using the cosine test formula:

\begin{equation}
    rot(T) = arccos \left(\frac{trace(R)-1}{2} \right)
\end{equation}

Despite the added complexity when compared to ATE, the RPE metric is more popular as it addresses the shortcomings of the ATE metric effectively, especially the accumulation of error.

\subsection{Average Translational and Rotational Error (Drift Rate)}
\label{subsec:driftrate}
\noindent
The Average Translational Error and Average Rotational Error are extensions of the RPE error metric and were popularized by the well-known KITTI dataset and benchmark \cite{KITTI}. The idea behind these metrics is to offer a measure of how much an algorithm drifts with distance. They are calculated by averaging the relative translation error and relative rotation error (as calculated in subsection \ref{subseq:rpe}) along every possible sub-trajectory of lengths (100, 200, 300, 400, 500, 600, 700, and 800) meters, the translation and rotation components for all lengths are then averaged and formatted as (translational error [\%] / rotational error [deg/100m]) as shown in Table \ref{tab:sota}. The Average Translational Error and Average Rotational Error are the most widely reported metrics for evaluating odometry.

\subsection{Other Metrics}
\noindent
An important factor in evaluating odometry algorithms is their time requirements, often referred to as the \textit{"run time"} of the algorithm. Odometry algorithms should be fully capable of real-time operation to ensure timely updates of the current pose estimate. This is contrary to what can be found in SLAM where parts of the algorithm (e.g., global bundle adjustment) can be performed either completely offline or on a separate process without time constraints. Comparing the time requirements of different odometry implementations is not an easy task; it is heavily dependent on the hardware used during testing. Another noteworthy metric is the so-called Spatial Cross Validation metric (SCV) which attempts to assess the generalization capabilities of learning-based approaches. The SCV is simply the mean of Average Translational Error and Average Rotational Error across sequences that have not been seen by the algorithm during the training phase. Finally, algorithms that estimate velocities instead of poses are typically evaluated by comparing reference and estimated linear velocity and angular velocity.

Readers interested in deeper analysis of odometry and SLAM evaluation metrics in addition to publicly available tools for evaluation can refer to \cite{KITTI,evo_tool,Scaramuzza_tutorial,what_should_be_learnt,kummerle2009metrics}.

\section{Radar Datasets}
\label{sec:radar_dataset}
\noindent
Datasets are the backbone of many computer vision and robotic-based research. Recording, curating, and labeling datasets is a lengthy endeavor that requires a large amount of resources. Fortunately, the common practice among researchers is to collect and share datasets and make them open access to the research community. This trend greatly helps make the most out of the collected datasets, increase the visibility of the research work, and accelerates the overall pace of research in the field. In this section, we summarize datasets that feature radar data, including both scanning and automotive radars. Datasets included here are usable in the context of odometry research, i.e., datasets containing radar data but were collected in a stationary setup (e.g., \textit{CARRADA} dataset \cite{carrada_dataset}) or datasets that do not include a source for ground truth pose information such as GPS/IMU, SLAM, or Visual Odometry (e.g., \textit{SCORP} dataset \cite{scorp_dataset}) are not applicable for odometry research and thus are not discussed here. Table \ref{tab:datasets} summarizes the datasets discussed in this section.

\textit{Oxford Radar RobotCar} \cite{oxford_radar_dataset}, 2019, Oxford, UK, is an extension of the original \textit{Oxford RobotCar} \cite{oxford_dataset} dataset. It is the most influential dataset in radar odometry research and the most used dataset for algorithm benchmarking. It offers a rich suite of sensors including four cameras, four lidars, one Navtech CTS350-X scanning radar, and GPS/INS module. Ground truth was generated offline by optimizing visual odometry, visual loop closures, and GPS/INS constraints. Despite its predominance, \textit{Oxford Radar RobotCar} falls short in terms of seasonal diversity and lacks sufficient snowy and foggy scenes.

\textit{Boreas} dataset \cite{boreas_dataset}, 2022, Toronto, Canada, is the most recent addition to the list. It is a multimodal, multi-seasonal, autonomous driving dataset that includes a monocular camera,  scanning radar (Navtech CIR304-H), and a 3D lidar (Velodyne Alpha-Prime with 128 beams). \textit{Boreas} dataset offers the greatest seasonal diversity to date with same-route traversals ranging from a perfectly sunny day to a very snowy storm. Its official website also has a leaderboard\footnote{https://www.boreas.utias.utoronto.ca/\#/leaderboard} for benchmarking odometry algorithms in $SE(2)$ and $SE(3)$ which makes it a valuable tool for odometry research.

\textit{RADIATE} dataset \cite{radiate_dataset}, 2019-2020, Edinburgh, UK, is another radar-centric autonomous driving dataset in winter weather conditions. It offers great seasonal diversity (snow, rain, fog, sun, overcast) and it was recorded at different times of the day (night and day). Its sensor suite includes a stereo camera (ZED stereo camera), a lidar (Velodyne HDL-32e), scanning Radar (Navtech CTS350-X), and GPS/IMU. One limiting factor of the \textit{RADIATE} dataset is the occasional loss of GPS signal in the ground truth data.

\textit{MulRan} dataset \cite{mulran_dataset}, 2020, Daejeon and Sejong, South Korea. \textit{MulRan} is focused on place recognition using radar and lidar and features multiple visits to some locations. It has a scanning radar (Navtech CIR204-H), a 3D lidar (Ouster OS1-64), and GPS data, but no camera recording. Ground truth was generated using SLAM-based procedure \cite{mulran_gt_slam} with the help of fiber optic gyro and Virtual Reference Station GPS. Limiting factors of \textit{MulRan} are the seasonal diversity and lack of visual data.

\textit{nuScenes} dataset \cite{nuscenes_dataset}, 2019, is another multimodal autonomous driving dataset with six visual cameras (Basler acA1600-60gc), five FMCW Automotive radars (Continental ARS 408-21), one lidar (Velodyne HDL32E), and IMU/GPS sensor. \textit{nuScenes} was recorded in two different cities: Boston, USA and Singapore which gives it great spatial diversity; however, it lacks the seasonal diversity that other datasets offer.

\textit{PixSet} dataset \cite{pixset_dataset}, 2020, Quebec and Montreal, Canada, consists of 97 sequences and 29 thousand annotated frames with bounding boxes which makes it more suitable for object detection and tracking research. However, since it also has IMU with RTK GPS sensor data, it should be usable for odometry and localization research. \textit{PixSet} is centered around flashing lidar sensor technology but also features an automotive-grade mmWave (radar TI AWR1843). The data was collected in urban environments mostly in daytime and rainy or cloudy weather conditions, it has four cameras, two lidars (one scanning and one flash lidar), and a GPS/IMU module.

The \textit{Marulan} dataset \cite{marulan_dataset}, 2010, features an FMCW radar that was built by the Australian Centre for Field Robotics (AFCR). It offers data from four 2D lidars, a visual camera, a thermal camera, an RTK GPS receiver, and an IMU sensor. limitations of \textit{Marulan} are that it was only recorded in off-road settings, slow driving speed, and the relatively slow scanning rate of the used radar.

\textit{RadarScenes} dataset \cite{radarscenes_dataset} 2016-2018, Ulm, Germany, was collected using four overlapping automotive radars. Its main focus is on object detection, tracking, classification, semantic and instance segmentation, and grid mapping using automotive radars. It has annotations for these perceptual tasks but also has GNSS information so it should be possible to use it for odometry research; however, it might not be very accurate.

Having multiple options for radar datasets is great for advancing research; nonetheless, the field of radar odometry lacks what KITTI \cite{KITTI} benchmark provides for vision-based research; a common and widely accepted reference for all researchers to benchmark their work.

\begin{table*}
\begin{center}
\caption{Publicly Available Datasets Suitable for Radar Odometry Research}
\label{tab:datasets}
\resizebox{\linewidth}{!}{%
\begin{tabular}{l|c|c|c|c|c}

\hline

\textbf{Dataset}  &    \textbf{Year}   &   \textbf{Radar Model}    &   \textbf{Other Sensors}     &   \textbf{Ground Truth}   &   \textbf{Seasonal Diversity}    \\
\hline
\textit{Oxford Radar RobotCar} \cite{oxford_radar_dataset} & 2019 & Navtech CTS350-X & four cameras, four lidars, GPS/INS   & Optimized VO  & \checkmark   \\

\textit{Boreas} dataset \cite{boreas_dataset}              & 2022 & Navtech CIR304-H & monocular camera, lidar, GPS/IMU     & GPS/IMU + RTX  & \checkmark \checkmark \checkmark \\

\textit{RADIATE} dataset \cite{radiate_dataset}            & 2019 & Navtech CTS350-X & stereo camera, lidar, GPS/IMU   & GPS/IMU  & \checkmark \checkmark \\

\textit{MulRan} dataset \cite{mulran_dataset}              & 2020 & Navtech CIR204-H & lidar, GPS                      & SLAM  & \checkmark \\

\textit{nuScenes} dataset \cite{nuscenes_dataset}          & 2019 & Continental ARS 408-21 & six cameras, lidar, GPS/IMU & GPS/IMU  & \checkmark \\

\textit{PixSet} dataset \cite{pixset_dataset}              & 2020 & TI AWR1843 & four cameras, two lidars (flash+scanning), GPS/IMU & GPS/IMU + RTK  & \checkmark \\

\textit{Marulan} dataset \cite{marulan_dataset}            & 2010 & Custom made FMCW scanning radar & camera, four 2D lidar, thermal camera, GPS/IMU & GPS/IMU + RTK  & \checkmark \checkmark \\

\textit{RadarScenes} dataset \cite{radarscenes_dataset}    & 2021 & 4 x Automotive Radars (unknown model) & Camera, GNSS    &   GNSS    &   \checkmark \\
\hline

\end{tabular}%
}
\end{center}
\end{table*}

\section{Radar Odometry}
\label{sec:radar_odometry}
\begin{figure}[!t]
\centering
\includegraphics[width=3.4in]{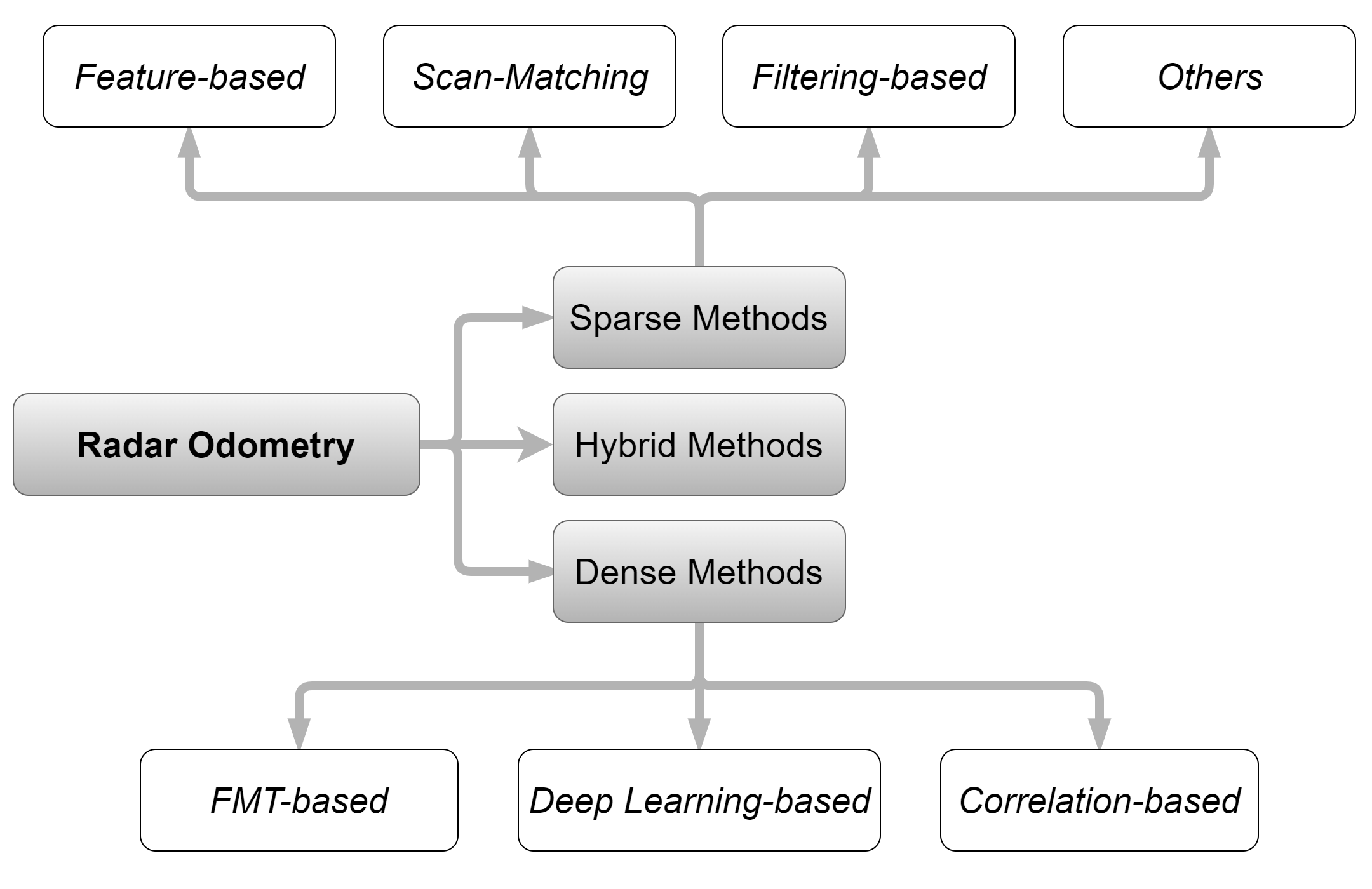}
\caption{The different branches of radar odometry methods based on input data representation, Sparse, Dense, and Hybrid. Sparse methods are further categorized into feature-based, scan-matching, filtering-based, and other methods. Dense methods are categorized as FMT-based, deep learning-based, and correlation-based.}
\label{fig:odometry_branches}
\end{figure}

\noindent
We follow the common trend in radar odometry literature and divide the approaches into three broad categories: \textit{Sparse Radar Odometry}, which relies on quantifiable number of detections (or landmarks) and use them for motion estimation, \textit{Dense Radar Odometry}, which processes whole scans in order to perform the said estimation, and \textit{Hybrid Radar Odometry} which relies on a combination of both. Fig. \ref{fig:sparse_vs_dense} shows the percentage of publications that are based on sparse, dense, and hybrid methods. It can be seen that sparse methods are much more popular than dense methods, this could be attributed to the higher computational requirements of dense methods as will be discussed later. We further divide sparse methods into feature-based, scan-matching, filtering-based, and others. Similarly, we divide dense methods into Fourier-Mellin Transform (FMT) -based, deep learning-based, and correlation-based. Fig. \ref{fig:odometry_branches} illustrates the described categorization of radar odometry methods. Following is a detailed discussion of each of these methods.

\subsection{Sparse Radar Odometry}
\label{subsec:sparse_radar_odometry}
\noindent
\subsubsection{Feature-Based}
Before going into the details of feature-based methods, it is worth highlighting that the words keypoints, features, and landmarks are used interchangeably in radar odometry literature, and they all mean what would the reader expect them to mean, salient properties of local segments within discrete one- or two-dimensional signals. We make a distinction between visual (or vision-based) features, which are features borrowed from the domain of visual cameras and are typically associated with image processing techniques, and special features, which were developed specifically for the goal of realizing radar odometry.

Beginning with visual features, Callmer \textit{et al.} \cite{callmer2011visualfeatures1} proposed a SLAM method for maritime applications using a spinning radar, the odometry step in their pipeline relies on detecting and matching Scale-Invariant Feature Transform (SIFT) features between consecutive scans. Using two sets of matched landmarks, they solve for change in \textit{x}, \textit{y}, and \textit{yaw} using least squares. They tackle the problem of motion distortion caused by the relatively slow scanning rate of their radar, which could cause significant shifts in landmarks' location, by dividing each scan into segments and estimating the change in pose incrementally between segments. In more recent work, Hong \textit{et al.} \cite{hong2020visualfeatures3} introduced a radar-based large-scale SLAM algorithm for autonomous driving with a focus on inclement weather. Their work is based on the scanning radar from Navtech. After converting the scans from polar coordinates to cartesian coordinates, they detect and match features using Speeded Up Robust Features (SURF) between the current frame and a keyframe. Outlier rejection and improved matching are achieved by employing pair-wise distance consistency constraints. Finally, they solve for the rigid body transform that aligns both sets of matched features using Singular Value Decomposition (SVD) and the well-known algorithm proposed in \cite{challis1995svd}. Hong \textit{et al.}, later on, improved their method in \cite{hong2024visualfeatures4} by replacing SURF features with a blob detector and Kanade–Lucas–Tomasi (KLT) tracker that tracks the detected features into the following scans instead of matching them. They solve for the aligning transformation between tracked features using the same approach (i.e., SVD), but they compensate for motion distortion introduced by the spinning radar using factor graph-based optimization. Barnes and Posner \cite{barnes2020visualfeatures2} leverage a U-net \cite{ronneberger2015unet} style Convolutional Neural Network (CNN) to learn robust features (or keypoints).  Crucial to their method, they use a differentiable pose estimator and a differentiable keypoints matcher allowing them to train the CNN to detect robust features for pose estimation. Their CNN architecture has two heads, one predicts the locations of keypoints and the other scores them, the cosine similarity function is used to give weights that indicate points with high scores and good matching scores. Finally, the pose is estimated using SVD. They use a loss function that combines both translational and rotational error using a factor ($\alpha$). Burnett \textit{et al.} \cite{motion_compensation} presented a method built on top of the well-known outlier rejection algorithm, the RANdom SAmple Consensus (RANSAC), Motion-Compensated RANSAC (MC-RANSAC) relies on extracting and matching Oriented FAST and Rotated BREIF (ORB) descriptors \cite{rublee2011orborigin} between consecutive cartesian radar scans, it then optimizes an objective function to estimate the velocity of the sensor taking into account the distortion caused by the ego-motion. The main goal of their work is to demonstrate the effect of motion distortion correction on odometry using slow-rotating sensors (e.g., Navtech scanning radars). They concluded that compensation for motion distortion had a marginal effect on odometry, around 10.0\% improvement, but a more significant improvement on localization was noticed. We postulate that studying the effect of motion distortion at higher driving speeds would lead to stronger justification for performing motion distortion. Burnett \textit{et al.} also proposed HERO \cite{burnett2021visualfeatures5}, short for Hybrid-Estimate Radar Odometry, a scanning radar odometry approach that combines unsupervised deep learning and probabilistic estimation. It builds on the authors' previous work on Exactly Sparse Gaussian Variational Inference (ESGVI) parameter learning framework \cite{barfoot2020ESGVI} for probabilistic state estimation, in addition to the work done by Barnes and Posner \cite{barnes2020visualfeatures2} on a U-net style CNN for feature detection using scanning radar data projected into cartesian coordinates, they tested their method on \textit{Oxford} and \textit{Boreas} datasets and reported the results as in Table \ref{tab:sota} for reference. Finally, Lim \textit{et al.} \cite{lim2023visualfeatures6} developed ORORA (abbreviation of Outlier-RObust RAdar odometry), a decoupled radar odometry method based on scanning radars. They build on the work in \cite{motion_compensation} for feature extraction and matching and add a graph-based outlier rejection step (i.e., the maximum clique of a graph). Finally, they estimate rotation and then translation in two separate steps. They tested their method using \textit{MulRan} dataset and reported an improved performance when compared to MC-RANSAC \cite{motion_compensation} mentioned earlier.

Relevant publications on non-visual features include the seminal work by Cen and Newman \cite{cen2018specialfeatures1}, based on Navtech spinning radar, they process each 1D radial signal (i.e., signals associated with each azimuth) and return useful landmarks based on statistical properties of the power return signal. Finally, they estimate the relative motion between the two sets of matched landmarks using SVD. The same authors later improved their approach in \cite{cen2019specialfeatures2}, where they expanded on their landmark extraction and association algorithms. Aldera \textit{et al.} \cite{aldera2019specialfeatures3} build upon these methods by leveraging deep learning for improved feature extraction. They used a U-Net style CNN \cite{ronneberger2015unet} that was weakly supervised using ground truth signal from visual odometry and trained it to filter radar returns to either "keep" or "reject" before using the scan for motion estimation as in \cite{cen2018specialfeatures1}. Similarly, Aldera \textit{et al.} \cite{aldera2019specialfeatures4} further built on \cite{cen2018specialfeatures1} and proposed a self-checking algorithm, this was achieved by first noting that the pair-wise compatibility matrix of associated landmarks has eigenvectors of similar magnitudes for well-matched landmarks, they then used a Support Vector Machine (SVM) that is weakly supervised by information from GPS as ground truth to distinguish between good and bad odometry estimates based on these eigenvectors. In a different work, Aldera \textit{et al.} \cite{aldera2022specialfeatures5} draw inspiration from previous work in visual odometry \cite{scaramuzza20091point} exploiting motion constraints on mobile robots to filter out outliers and use a single landmark association to solve for ego-motion estimation based on the constant curvature motion assumption. Finally, Zhu \textit{et al.} \cite{zhu2023specialfeatures6} leveraged deep learning and proposed a model named DeepEgo based on automotive radars to achieve instantaneous estimation. Hence, eliminating the need for scan-to-scan processing. Trained and tested on \textit{RadarScenes} dataset, their proposed model takes radar point clouds and outputs point-wise weights and point-wise offsets for which weighted least square is used to estimate the ego-motion of the vehicle. Their model is specific for each radar, i.e., in the case of utilizing multiple radars for odometry, multiple instances of the model have to be trained and deployed, moreover, how well this approach generalizes over other datasets or sensor types was not clarified in their work. 

\subsubsection{Scan-Matching} In general, the idea of scan-matching approaches is to find the rigid body transformation that \textit{best} aligns two preprocessed radar scans. The main difference between scan-matching approaches and feature-based approaches is the lack of association or correspondence in scan-matching (i.e., points in the source scan are not matched with points in the target scan). Scan-matching methods are usually iterative in nature and attempt to refine the transformation with each iteration to minimize an error metric between the aligned scans. We distinguish between three techniques commonly used in scan-matching methods: Iterative Closest Point (ICP), Point to Normal Matching (P2N), and the Normal Distribution Transform (NDT). These three subcategories are sometimes referred to by the literature as point-to-point, point-to-line, and point-to-distribution matching, respectively.

Originally developed for 3D point cloud registration \cite{besl1992icporigin}, Iterative Closest Point (ICP) has been very successful in lidar-based odometry and has seen numerous improvements in recent decades. Naturally, this success in the lidar domain has been extrapolated to the radar domain. Marck \textit{et al.} \cite{marck2013icp1} presented a SLAM algorithm using a spinning radar from TNO. The ego-motion estimation step in their pipeline uses vanilla ICP to estimate rotation and translation between two radar scans. Their approach was demonstrated in an indoor environment. Fritsche and Wagner \cite{fritsche2017icp2} presented another SLAM algorithm based on lidar/radar fusion, the radar used is a spinning radar (called MPR, for Mechanical Pivoting Radar) developed by Fraunhofer Institute for High Frequency Physics and Radar Techniques (FHR). In their approach, they used hand-made heuristics to combine lidar and radar scans and feed these scans to ICP in order to obtain the transformations. Their fusion scheme was assessed in a visually challenging indoor environment qualitatively. 

The term Point-to-Normal Matching (P2N) is used in this article to group the work presented by Adolfsson \textit{et al.} in \cite{adolfsson2021p2n1,adolfsson2022p2n2} and Alhashimi \textit{et al.} in \cite{alhashimi2021p2n3} which are based on point to line matching strategy. CFEAR approach (which stands for Conservative Filtering for Efficient and Accurate Radar odometry) \cite{adolfsson2021p2n1,adolfsson2022p2n2} begins with filtering radar scans of a spinning radar (from Navtech) by keeping only the $k$ strongest returns, where $k$ is a tunable parameter. The kept points are then clustered together based on their statistical properties, i.e., mean and covariance, and treated as surfaces for which the normals are calculated using the eigenvectors of the covariance matrix. Scan-to-keyframe(s) matching is then performed to align the current scan with the latest keyframe or multiple keyframes by finding a transformation that minimizes a point-to-line cost function. They tested their method on several publicly available datasets, \textit{Oxford} and \textit{MulRan} datasets, and reported a remarkably good accuracy as can be seen in Table \ref{tab:sota}. Their approach was further improved in \cite{alhashimi2021p2n3} by replacing the filtering technique (i.e., CFEAR) with BFAR filtering,  where BFAR (Bounded False Alarm Rate) is a generalization of the classical Constant False Alarm Rate (CFAR) detector and uses an affine transformation with two learnable parameters for improved detection.  

The Normal Distribution Transform (NDT) was first proposed by Biber \textit{et al.} \cite{biber2003ndtorigin} for point cloud registration. Just like ICP algorithms, it gained popularity in lidar-based applications and soon got ported to radar-based ones. Vanilla NDT divides the first point cloud into cells and represents points in each cell by a Gaussian normal distribution. It then aims to find a transformation that maps points in the second point cloud to the first point cloud by maximizing the likelihood that points at given spatial coordinates are drawn from the corresponding distribution. Rapp \textit{et al.} \cite{rapp2015ndt1} presented an ego-motion estimation method using both spatial and Doppler information from one or multiple automotive radars with sparse detection. The method begins by representing radar scans as normal distributions by using \textit{k-means} clustering, then finds the transform that maximizes the geometric overlap between previous detection and new detections, they also use Doppler information to weigh the distributions and suppress dynamic objects in the scene. Rapp \textit{et al.} later improved this work in \cite{rapp2017ndt2} by testing more sophisticated clustering methods: Density Based Spatial Clustering of Applications with Noise (DBSCAN), Ordering Points To Identify the Clustering Structure (OPTICS), in addition to \textit{k-means} from their previous work. They concluded that DBSCAN is the best clustering method for this application. Work presented by Kung \textit{et al.} \cite{kung2021ndt3} is the only approach targeting both automotive and scanning radar sensors. Their simple yet efficient pipeline consists of fixed thresholding of radar scans, generation of probabilistic radar submap of sparse Gaussian Mixture Models, and applying the NDT to find the transform that best aligns two consecutive scans. They tested their method on both \textit{Oxford} dataset (scanning radar) and \textit{nuScenes} dataset (automotive radar) and achieved results comparable to the state-of-the-art (see Table \ref{tab:sota}) despite its simplicity. Almalioglu \textit{et al.} \cite{almalioglu2021ndt4} also employed NDT to realize ego-motion estimation for indoor environments using automotive radar. They improve the performance of the NDT by fusing data from an IMU after the scan matching step using Unscented Kalman Filter (UKF) and utilizing a Recurrent Neural Network (RNN) model to act as a motion model and benefit from previous pose estimates. Zhang \textit{et al.} \cite{zhang2023ndt5} further extended NDT applications in radar odometry with scanning radar to include radar-based localization, they present a specialized denoising technique for Navtech radars and an improved NDT scan matching based on adjustable outlier ratio rather than weighted distributions.  They test their approach on three public datasets: \textit{Oxford}, \textit{MulRan}, and \textit{RADIATE} datasets and show that their method surpasses state-of-the-art in radar odometry (see Table \ref{tab:sota}). Finally, Haggag \textit{et al.} \cite{haggag2022ndt6} also used NDT to tackle radar odometry using automotive radar. Their work is more oriented toward the point registration problem in an uncertainty-aware fashion with the goal of using the covariance information to extend the method and fuse data from other sensors.

NDT-based approaches are considered fast and effective, they efficiently break down huge point clouds into manageable piece-wise normal distributions. They are, however, similar to ICP-based methods; iterative in nature and could be sensitive to the initial configuration.

\subsubsection{Filtering-Based}

We refer to a class of methods that relies on Kalman Filter (KF) or one of its variants as the main estimator as filtering-based methods. Holder \textit{et al.} \cite{holder2019filter1} proposed a graph-based SLAM approach based on an automotive radar with sparse returns. Although they use ICP scan matching in the localization and mapping steps, their odometry relies on UKF with a motion model and a measurement model. They first estimate the ego-velocity using the method proposed by \cite{kellner2013kinej1}, then separate static and dynamic objects in the scene. Finally, they fuse data from the vehicle's Controller Area Network (CAN) bus, including gyroscope, wheel encoders, and steering wheel, to estimate the states of the vehicle. In a similar work, Liang \textit{et al.} \cite{liang2020filter2} proposed a method based on Error-State Extended Kalman Filter (ESEKF), they fused data from automotive radar with IMU, GNSS, lidar, and camera to achieve overall improved robustness and redundancy of their system. Finally, Araujo \textit{et al.} \cite{araujo2023filter3} also fused IMU and radar using a Kalman Filter variant, the Invariant Extended Kalman filter (IEKF). They also utilized a deep neural network model to clean radar data as a preprocessing step before feeding it to the IEKF for velocity estimation.

In general, filtering-based approaches are well-suited for fusing data from several sensors, they harness the complementary nature of multiple sensors to obtain more accurate and robust performance. They, however, don't generalize well across platforms as they are considered model-reliant (i.e., motion model and measurement model), they require parameter tuning, and if not carefully implemented, can be computationally demanding.

\subsubsection{Others}
This last subcategory in sparse methods is used to group \textit{other} special methods that do not exactly fit in any of the groups discussed previously. These methods are based on unique kinematic, geometric, or statistical formulations.

Ng \textit{et al.} \cite{ng2021kinex1} developed a continuous-time method for odometry using multiple automotive radars and an IMU sensor.  They model the trajectory as cubic B-splines and use a sliding window optimization framework to obtain the trajectory parameters. They argue that the continuous representation of the trajectory and the enforced smoothness are beneficial for finding poses and their derivatives at any point in time, even at instants where no information is available from any sensor. They compare their approach with a discrete-time implementation and show that their approach offers improved translational error but suffers from a marginal increase in rotational error.  

Ghabcheloo and Siddiqui \cite{ghabcheloo2018kinex2} propose a purely kinematic-based approach where the positions of stationary objects and the vehicle are described with respect to a reference frame, they then derive expressions for linear and angular velocities which they conclude that are not solvable using a single radar and use information from a gyro to overcome this problem. The algorithm estimates the velocities using a pair of points and uses RANSAC to reject outliers. Despite the plausibility of the formulation; their method is very sensitive to the ratio of static to dynamic objects in the scene. 

Scannapieco \cite{scannapieco2019kinex3} presented an outlier rejection scheme for automotive radar-based odometry. Their statistical approach relies on classifying returns as static and dynamic based on their bearing and radial velocity. The main assumption is that static targets will cluster around the median of all measurements and detections that are distant from the median are considered outliers and rejected. They tested this technique in both actual and simulated environments and showed successful outlier rejection results.

Kellner \textit{et al.} \cite{kellner2013kinej1} proposed a method to estimate the ego-velocity of a vehicle using the radial velocity measurements of single or multiple automotive radars. Critical to their approach, a separation step using RANSAC classifies objects in the scene into stationary and dynamic, assuming that the majority of the objects are stationary, they analyze the velocity profiles of stationary objects to infer the velocity of the vehicle itself using the Ackerman condition. Kellner \textit{et al.} \cite{kellner2014kinej2} later expand on this method by calculating the uncertainty of the velocity estimates. The advantages of this approach include the fact that it is scalable to multiple radars and that the velocity is estimated in a single measurement cycle. However, their method has been tested at relatively low driving speeds and has limited capabilities when there are many dynamic objects in the scene. Barjenbruch \textit{et al.} \cite{barjenbruch2015kinej3} combined spatial and Doppler information to tackle the radar odometry problem. Based on automotive radar, their proposed method reprojects radar scans into cartesian coordinates and represents measurements as Mixture of Gaussians, it later adds Doppler velocity estimate to jointly optimize for ego-motion. They claim that their method is scalable to more than one radar sensor and their tests demonstrate improved estimation accuracy when compared to \cite{kellner2013kinej1}. Zeng \textit{et al.} \cite{zeng2023kinej4} investigated the issue of velocity ambiguity in Doppler radar measurements and argued that addressing this problem would significantly improve the accuracy of radar-based odometry. They proposed a method to jointly estimate velocity ambiguity and ego-motion utilizing a novel clustering technique. They compared their approach with other instantaneous ego-motion estimation methods including the work of Kellner \textit{et al.} \cite{kellner2013kinej1} and reported improved overall performance.

Vivet \textit{et al.} \cite{vivet2012kinen1, vivet2013kinen2} took a completely different approach to the problem; where the majority of work based on spinning radars compensate for motion distortion (see section \ref{sec:radar_sensor}), Vivet \textit{et al.} tried to use this information to estimate the states of the moving platform. Their method relies on the extraction of landmarks using CFAR and finding associations of these landmarks between two consecutive scans, these are used along with a formulation for motion distortion to estimate the ego-velocity. Velocity estimates are also sent to an Extended Kalman Filter (EKF) and RANSAC for smoothing and outlier rejection. Despite its novelty, it is unclear how accurate this method would perform in very low-speed or high-speed scenarios or in the case of aggressive turns. The method is also restricted by the assumption that the majority of surrounding objects are static and it can be very sensitive to the quality of landmark detection and association step. Finally, Retan \textit{et al.} presented an odometry method using automotive radars based on constant velocity \cite{retan2021kinel1} and constant acceleration \cite{retan2022kinel2} motion models on $SE(3)$, they proposed a sliding window optimization framework and data association technique and compared their results with scanning radar-based methods.

Contrasting these methods is not a straightforward task; some methods were designed for scanning radars, others were designed for automotive radars. Many researchers tested their methods using unpublished datasets that they have collected with their unique sensor setup, driving environment, and mobile platform. These factors hinder a useful quantitative comparison between these methods. There are, however, some researchers who used the \textit{Oxford} dataset to benchmark their scanning radar-based methods, published results on \textit{Oxford} dataset and evaluation sources are collected in Table \ref{tab:sota}.

\subsection{Dense Radar Odometry}
\label{subsec:dense_radar_odometry}
\noindent
Dense approaches draw inspiration from \textit{direct methods} in visual odometry where optical flow is used to estimate pose changes. In the context of radar odometry, dense methods are characterized by taking full radar scans as inputs with minimal to no preprocessing. Dense methods can be divided into FMT-based, deep learning-based, and correlation-based.
\subsubsection{Fourier-Mellin Transform (FMT) based}
 One of the earliest examples of dense radar odometry is the work of Checchin \textit{et al.} \cite{checchin2010fmt} where an FMCW panoramic K-band scanning radar called K2Pi was used to generate dense 2D images of the surrounding environment. They used the FMT, a well-known image registration technique from computer vision, to calculate the rotation and translation between two consecutive dense radar scans. In brief, their algorithm starts with thresholding and converting the images into polar coordinates. Next, they use the Fast Fourier Transform (FFT) in order to transform the image into the frequency domain and the normalized cross-correlation to find the rotation between the two images. Once the original image is corrected for rotation, the FFT and normalized cross-correlation steps are repeated to calculate the shifts in the $x$ and $y$ dimensions. A very similar approach was presented by Park \textit{et al.} \cite{park2020pharao} which also uses an FMCW scanning radar from Navtech and applies FMT in two steps, the first step on a downsampled scan that retrieves rotation and an initial estimate of translation. The second step refines the translation using the full-resolution image. Additionally, they further improve the odometry results by adding keyframe selection and local graph optimization steps to their pipeline. 
 \subsubsection{Deep Learning-based}
Barnes \textit{et al.} \cite{barnes2020mbym} used an FMCW scanning radar from Navtech and leveraged a U-net style CNN to generate masks that suppress noise and artifacts in the radar scan. The CNN was trained in a self-supervised manner, i.e., ground truth pose information from visual odometry was used to train the masking network. The cross-correlation values between two consecutive filtered scans at all possible rotations were calculated and the maximum cross-correlation is used to estimate the pose. This approach was later improved by Weston \textit{et al.} \cite{weston2022fastmbym} where they decoupled translation and rotation problems into two steps and utilized the translation invariance property of the Fourier Transform on polar coordinates to avoid the exhaustive search for maximum cross-correlation. The work of Lu \textit{et al.} \cite{lu2020milliego} is the only case of an automotive radar, the Texas Instruments TI AWR1843, being used in a dense radar odometry method. It is also the only example of a complete end-to-end deep learning-based radar odometry pipeline. In their work, they solve the sparsity problem of automotive radars by reprojecting it into a 2D image-like representation. These images are then fed to a CNN while IMU data is fed to an RNN. The outputs of both networks are fused together using a mixed attention mechanism and then forwarded to a Long Short-Term Memory network (LSTM) to benefit from previous poses in estimating the current pose. Finally, a fully connected network is used to regress the pose.
\subsubsection{Correlation-based}
A completely different perspective was presented by Ort \textit{et al.} \cite{ort2020gpr} where a Ground Penetrating Radar (GPR) was used for \textit{localization} under inclement weather. GPRs are typically used for archeological, geographical, and construction-related applications such as underground mapping, soil classification, pipes and mines detection, and many others. Their use in robotic applications, however, is quite unique. Although their work is considered a localization method, it should be possible to repurpose it as an odometry method. They used a custom-designed radar by MIT Lincoln Laboratory and performed localization by finding the maximum correlation of the current scan with a database of pre-stored scans. The accuracy is further improved by fusing data from wheel encoders and an IMU using EKF. 

Dense approaches are intuitive and easy to implement. They can be, however, computationally expensive due to the relatively larger volume of data being processed. All methods discussed in dense methods are formulated in $SE(2)$ following the 2D nature of the scanning sensors, this is usually acceptable for automotive applications; however, there could be some scenarios where $SE(2)$ is not enough in applications where elevation, pitch, and roll are observable and of certain importance to the task. Additionally, motion distortion is often neglected in dense approaches despite its potentially significant bad influence on the results. We also highlight the fact that dense approaches are more suitable for scanning radars that generate dense scans unless sparse points from other types of radars are converted into other representations similar to \cite{lu2020milliego} which adds another preprocessing step. GPR radars have great potentials as one of the most resilient sensors to weather and environmental changes, they are, however, still bulky and not very suitable for automotive applications and more work is needed on GPR-based approaches. Deep learning-based methods typically face the challenges of generalization over unseen data, new environments, or different sensor specifications, moreover, the loss function used in \cite{lu2020milliego} is the same for both error sources (i.e., translation and rotation), we postulate that using multitasking neural networks to estimate translation and rotation separately might improve the accuracy of both, alternatively, a model architecture with two parallel pipelines could be used to estimate the two components of pose separately.

\subsection{Hybrid Radar Odometry}
\label{subsec:hybrid_radar_odometry}
\noindent
Hybrid methods refer to approaches that use both radar data representations discussed previously (i.e., dense and sparse) regardless of the type of used radar. The only example that falls into this category is the work of Monaco \textit{et al.} \cite{monaco2020radarodo} where they decouple the problem into two sub-problems: translation estimation and rotation estimation. For translation estimation, they use the sparse points to estimate the velocity by solving the equation of motion using least squares, they then use RANSAC to remove dynamic objects from the scene. For rotation estimation, they first translate the images, based on the previous step, and then look for a maximum correlation between consecutive images in order to estimate rotational velocity.  Their work requires information that is not typically provided by automotive radars so they resolve to use a scanning radar (Delphi ESR 2.5) and emulate sparse points from its scans. While decoupling the problem into two sounds like a good idea following a "divide and conquer" strategy for pose estimation, the advantages of combining dense and sparse representations are not very clear in their work.

\begin{table*}
    \begin{center}
    
    \caption{Performance Of The State-of-the-Art Scanning Radar Odometry Methods On Eight Sequences from \textit{Oxford Radar RobotCar} Dataset}
    \label{tab:sota}
    \resizebox{\linewidth}{!}{%
    \begin{tabular}{c|l|c|cccccccc|c|c}

        \hline
        &      &            & \multicolumn{8}{c|}{\textbf{Sequence}} &      &          \\
        \textbf{Sensor} & \textbf{Method} & \textbf{Evaluation} & \textbf{10-12-32} & \textbf{16-13-09} & \textbf{17-13-26} & \textbf{18-14-14} & \textbf{18-15-20} & \textbf{10-11-46} & \textbf{16-11-53} & \textbf{18-14-46} & Mean & Mean SCV \\ \hline \hline

             & Visual Odometry\cite{churchill2012vo}        &\cite{barnes2020mbym}              &   NA          &   NA          &   NA          &   NA          &   NA&NA&NA&NA&3.98/1.0& 3.98/1.0   \\
     Camera  & ORB-SLAM2$^\ast$ \cite{mur-artal2017orbslam}         &\cite{hong2024visualfeatures4}     &   6.09/1.6    &   7.03/2.0    &   6.41/1.7    &   7.05/1.8    &   11.5/3.3&6.11/1.7&6.16/1.7&7.17/1.9&7.01/1.9& 7.01/1.9   \\  \hline
     Lidar & SuMa$^{\ast\ast}$\cite{behley2018suma}                   &\cite{hong2024visualfeatures4}     &   1.1/0.3     &   1.2/0.4     &   1.1/0.3     &   0.9/0.1     &   1.0/0.2&1.1/0.3&0.9/0.3&1.0/0.1&1.03/0.3& 1.03/0.3  \\      \hline
             & Cen RO\cite{cen2018specialfeatures1}         &\cite{barnes2020mbym}              &   NA          &   NA          &   NA          &   NA          &   NA&NA&NA&NA&3.72/0.95&3.63/.96    \\
             & Under the Radar\cite{barnes2020visualfeatures2}&\cite{barnes2020visualfeatures2} &   NA          &   NA          &   NA          &   NA          &   NA&NA&NA&NA&2.05/0.67& NA \\
             & Hong Odometry\cite{hong2024visualfeatures4}  &\cite{hong2024visualfeatures4}     &   2.32/0.70   &   2.62/0.7    &   2.27/0.6    &   2.29/0.7    &   2.25/0.7&2.16/0.6&2.49/0.7&2.12/0.6&2.24/0.7&  2.24/0.7  \\
             & Hong Radar SLAM$^\ast$\cite{hong2024visualfeatures4}&\cite{hong2024visualfeatures4}     &   1.98/0.60   &   1.48/0.5    &   1.71/0.5    &   2.22/0.7    &   1.77/0.6&1.96/0.7&1.81/0.6&1.68/0.5&1.84/0.6& 1.84/0.6 \\
             & Kung RO\cite{kung2021ndt3}                   &\cite{kung2021ndt3}                &   NA          &   NA          &   NA          &   NA          &   NA&NA&NA&NA&1.96/0.6&1.96/0.6 \\
             & Hero\cite{burnett2021visualfeatures5}        &\cite{burnett2021visualfeatures5}  &   1.77/0.62   &   1.75/0.59   &   2.04/0.73   &   1.83/0.61   &   2.20/0.77&2.14/0.71&2.01/0.61&1.97/0.65&1.96/0.66&NA \\
     Radar   & MC-RANSAC\cite{motion_compensation}          &\cite{burnett2021visualfeatures5}  &   NA          &   NA          &   NA          &   NA          &   NA&NA&NA&NA&3.32/1.09&  NA \\
             & MbyM (Dual Cart.)\cite{barnes2020mbym}       &\cite{barnes2020mbym}              &   NA          &   NA          &   NA          &   NA          &   NA&NA&NA&NA&1.16/0.3& 2.78/0.85  \\
             & fast-MbyM\cite{weston2022fastmbym}           &\cite{weston2022fastmbym}          &   NA          &   NA          &   NA          &   NA          &   NA&NA&NA&NA&2.06/0.60& NA \\
             & CFEAR\cite{adolfsson2021p2n1}                &\cite{adolfsson2021p2n1}           &   1.64/0.48   &   1.86/0.52   &   1.66/0.48   &   1.71/0.49   &   1.75/0.51&1.65/0.48&1.99/0.53&1.79/0.5&1.76/0.50&1.76/0.50    \\
             & CFEAR-2\cite{adolfsson2022p2n2}              &\cite{adolfsson2022p2n2}           &   \textbf{1.05/0.34}   &   \textbf{1.08/0.34}   &   \textbf{1.07/0.36}   &   \textbf{1.11/0.37}   &   \textbf{1.03/0.37}&\textbf{1.05/0.36}&\textbf{1.18/0.36}&\textbf{1.11/0.36}&\textbf{1.09/0.36}& NA \\
             & BFAR\cite{alhashimi2021p2n3}                 &\cite{alhashimi2021p2n3}           &   1.48/0.42   &   1.54/0.43   &   1.52/0.44   &   1.49/0.45   &   1.45/0.44&1.54/0.45&1.65/0.49&1.65/0.50&1.55/0.46 &1.55/0.46   \\
             & SDRO\cite{zhang2023ndt5}                     &\cite{zhang2023ndt5}               &   1.43/0.32   &   1.58/0.24   &   1.52/0.27   &   1.58/0.35   &   1.66/0.24&1.63/0.32&1.78/0.34&1.68/0.38&1.65/0.30& 1.65/0.30 \\ \hline
             \multicolumn{13}{l}{Results reported here were collected from several sources for easier comparison between methods. NA indicates results are not available from the source or not applicable.}\\
             \multicolumn{13}{l}{ Standard KITTI evaluation metrics are used here; translational error [\%] / rotational error [deg/100m]. $^\ast$ method uses loop closure. $^{\ast\ast}$ method failed to complete the trajectory.}\\
            \end{tabular}%
    }

\end{center}
\end{table*}

\section{Radar Odometry and Sensor Fusion}
\label{sec:fusion-based_radar_odometry}
\noindent
Sensors degrade variably in different challenging environments. Their advantages and disadvantages are considered to be complementary in nature, therefore, fusing multimodal data offers a plausible solution to overcome the limitations of each sensor individually \cite{wang2020fusion1,jahromi2019fusion2,rawashdeh2022drivable}. Moreover, most of the sensors discussed here are already widely adopted in the domain of autonomous vehicles, thus, it makes sense to develop algorithms that make use of all hardware and sensing modalities available onboard. Examples of problems associated with radar sensors include ghost objects, low resolution, multipath reflections, motion distortion, and saturation. Cameras have their own problems too, for example, sensitivity to lighting and weather conditions. Lidars are also affected, to a lesser extent, by adverse weather conditions and motion distortion. Typical IMUs are noisy and drift quickly, and finally, wheel encoders are vulnerable to wheel slip.
In addition to various weaknesses associated with these sensors, unimodal odometry algorithms usually have their own inherent problems as well. For example, monocular visual odometry can estimate motion up to a scale, stereo visual odometry is very sensitive to the quality of the calibration, rectification, disparity, and triangulation, it also degrades to monocular when the depth is much larger than the baseline of the stereo pair. Radar odometry methods based on scanning radars usually have low frequencies. Lidar odometry methods are computationally demanding in general and their scan-matching based methods require good initialization.
Despite all the anticipated merits of using sensor fusion in odometry, we highlight the fact that the percentage of published work on radar odometry in which a sensor fusion technique was employed is somewhat less than expected (Fig. \ref{fig:fusion_vs_nofusion}), we hypothesize that the reasons behind this are the lack of standardization in hardware and methods used for sensor fusion, in addition to the maturity and robustness of unimodal methods that make justifying multimodal approaches increasingly difficult.

Following is a brief overview of sensor fusion methods found in radar odometry literature. We restrict the discussion here to methods in which data from other sensors (e.g., camera, IMU) should be available at runtime, methods that used data from other sensors for training, testing, calibration, or as a source of ground truth only are not considered fusion-based methods. The most popular sensor configuration used for radar odometry is a radar and an IMU; the high sampling rate of a typical IMU complements that of the radar while the radar periodically corrects the undesirable drifts of the IMU. This sensor duo has found great success in UAVs, where a Kalman filter or one of its variants is used to fuse data from both sensors \cite{quist2016uav_1,mostafa2018uav_2,kramer2020uav_3,doer2020uav_4,doer2021uav_5,doer2022uav_6}. Similarly, work presented by Almalioglu \textit{et al.} \cite{almalioglu2021ndt4} and Araujo \textit{et al.} \cite{araujo2023filter3} are based on using a Kalman filter variant to fuse radar and IMU data. One advantage of fusing with Kalman filter is the relative ease of extending the formulation to include more sensors, for example, Ort \textit{et al.} \cite{ort2020gpr} fused data from a ground penetrating radar, wheel encoders, and an IMU. Holder \textit{et al.} \cite{holder2019filter1} fused preprocessed radar data, gyroscope, wheel encoders, and steering angle. Lastly, Liang \textit{et al.} \cite{liang2020filter2} fused radar, lidar, camera, GNSS, and IMU in one scalable framework. On the other hand, Lu \textit{et al.} \cite{lu2020milliego} used deep learning and proposed a mixed-attention mechanism of self- and cross-attention to fuse data from radar and IMU, they claim that their method outperforms simple features concatenation inside the model and that their model can be easily scaled to include more sensors. Fritsche and Wagner \cite{fritsche2017icp2} used hand-crafted heuristics to combine detections from a radar and a lidar into one unified scan. Ng \textit{et al.} \cite{ng2021kinex1} were able to combine asynchronous data from a radar and an IMU using their proposed continuous-time formulation. Finally, Ghabcheloo and Siddiqui \cite{ghabcheloo2018kinex2} used data from a gyroscope to only circumvent a singularity case in their formulation.

\section{Radar Odometry and Machine Learning}
\label{sec:radar_odometry_and_deep_learning}
\noindent
Machine learning techniques have been widely used in various robotic perception tasks and have been successfully employed in countless examples in autonomous vehicles and robotics in general \cite{grigorescu2020dlsurvey,Abu-Alrub2022weather}. Learning-based techniques harness the power of modern-day hardware to process large amounts of data in order to leverage modeling-free methods. Nevertheless, we also find the number of publications that make use of any learning-based techniques is less than expected (Fig. \ref{fig:learning_vs_nolearning}). This \textit{might} be related to the difficulty associated with generalizing learning-based methods, a problem that seems to be common in the machine learning domain but particularly harder to tackle in odometry and SLAM.

Following is a brief overview of leaning-based methods found in radar odometry literature. The work presented by Barnes \textit{et al.} \cite{barnes2020mbym} and Weston \textit{et al.} \cite{weston2022fastmbym} is based on training a CNN to predict masks that can be used to filter out noisy radar scans. It uses visual odometry as a source of ground truth poses during the training phase. Aldera \textit{et al.} \cite{aldera2019specialfeatures3} also used a U-Net style CNN to generate masks that filter out radar noise and artifacts, they used visual odometry as ground truth. Barnes and Posner \cite{barnes2020visualfeatures2} used a U-Net style CNN to predict keypoints, scores, and discriptors. Burnett \textit{et al.} \cite{burnett2021visualfeatures5} also trained a U-Net CNN to predict keypoints, scores, and descriptors; however, they leveraged unsupervised learning to train their model. Aldera \textit{et al.} \cite{aldera2019specialfeatures4} used an SVM to classify eigenvectors of the compatibility matrix of associated landmarks in order to distinguish between good and bad estimates. Araujo \textit{et al.} \cite{araujo2023filter3} used a CNN as preprocessing step only to denoise radar data. Zhu \textit{et al.} \cite{zhu2023specialfeatures6} developed a Neural Network model that processes radar point cloud and generates point-wise weights and point-wise offsets that can be used in other stages of the algorithm for motion estimation. Almalioglu \textit{et al.} \cite{almalioglu2021ndt4} used an RNN to act as a motion model so that it can benefit from information in previous poses and better capture the dynamics of the motion. In Lu \textit{et al.} \cite{lu2020milliego}, various learning techniques were used, starting with a CNN to process radar data, next to an RNN to process IMU data, followed by a mixed-attention mechanism that is used to fuse both streams together, followed by an LSTM that benefits from previous poses before feeding its output to a fully connected network for pose regression. Finally, Alhashimi \textit{et al.} \cite{alhashimi2021p2n3} proposed a radar filter that is based on learnable parameters.

To this end, it seems that scanning radars are more common in learning-based techniques than automotive radars. This is probably due to the fact that scanning radars generate much richer data for which deep learning models become very appealing. Moreover, we notice that CNNs are the most popular learning technique in radar odometry, this is due to the resemblance between the scans of scanning radar and visual images where using CNNs for tasks like semantic segmentation and object detection is very common.

\section{Discussion, Challenges, and Future Recommendations}
\label{sec:problems&recommendaitons}
\begin{figure}[!t]
    \centering
    \subfloat[]{\includegraphics[width=3.4in]{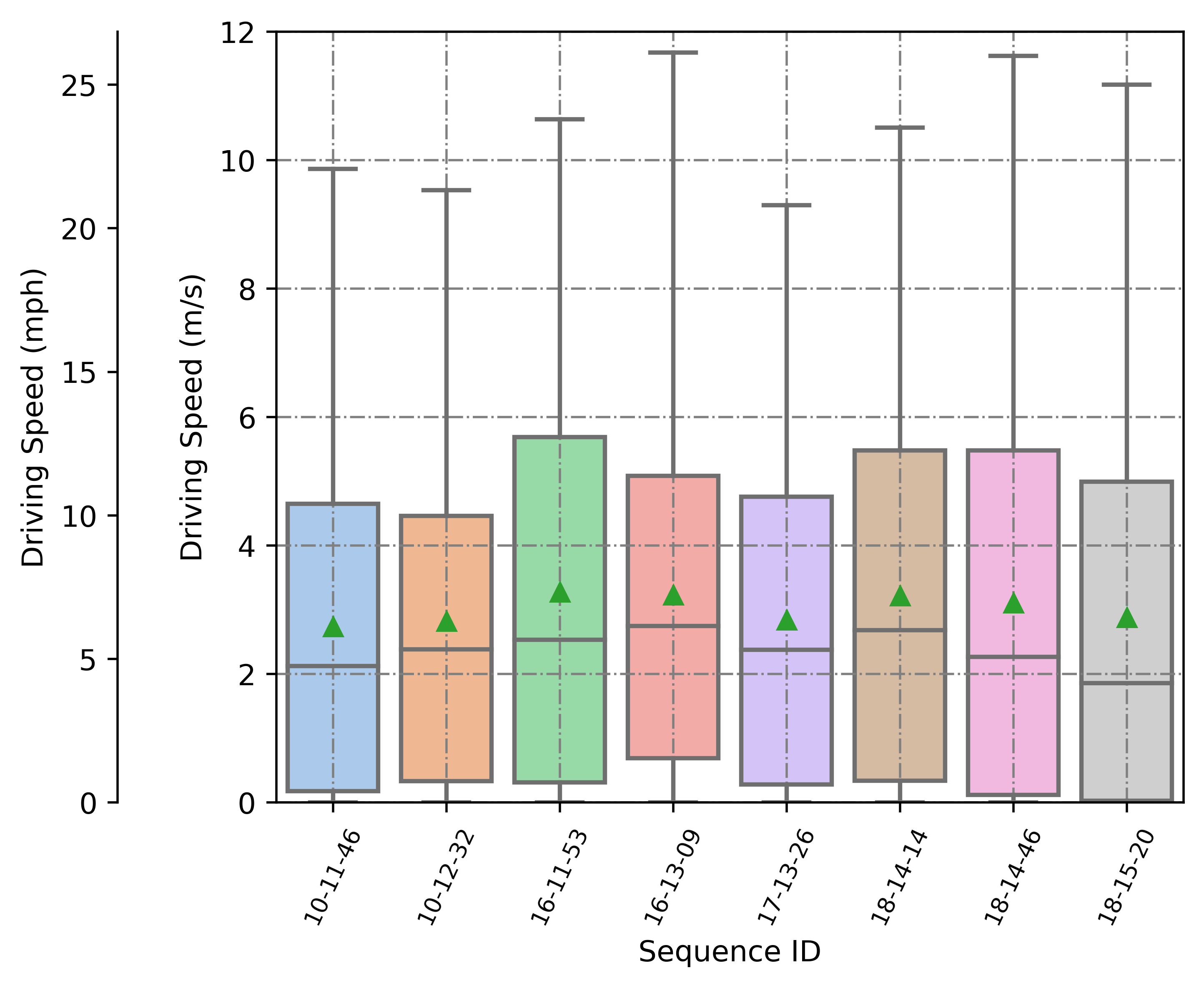}
    \label{fig:oxford_speed_analysis}}
    \hfil
    \subfloat[]{\includegraphics[width=3.4in]{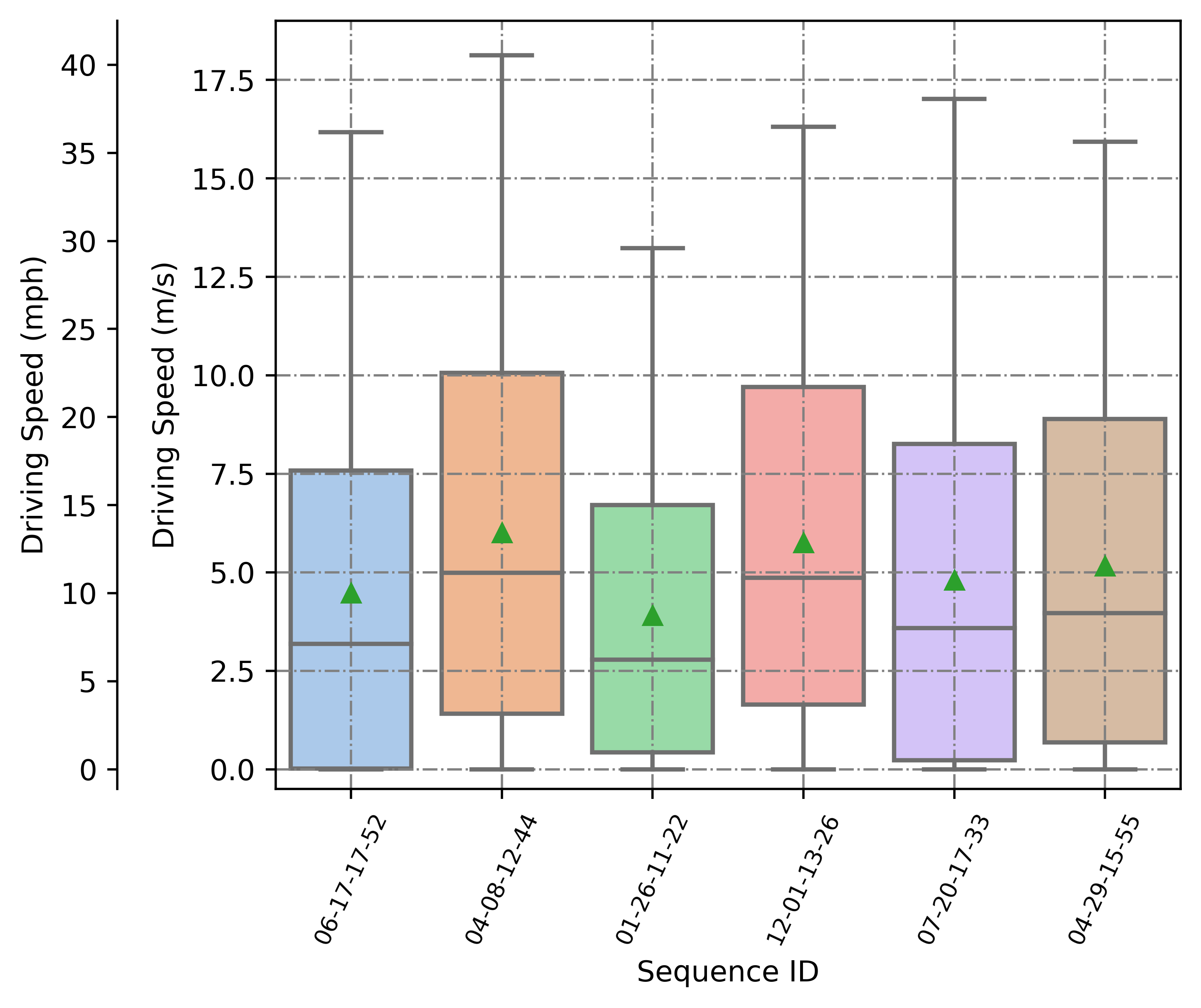}
    \label{fig:boreas_speed_analysis}}
    \hfil
    \caption{Statistics for the driving speed (in m/s and mph) while recording \textit{Oxford} and \textit{Boreas} datasets. Lines in the middle of the boxes indicate the medians. Green triangles indicate the averages. Speeds were calculated using ground truth data published with each dataset. (a) Boxplots of the driving speed while collecting \textit{Oxford} dataset for the eight most popular sequences which also appear in Table \ref{tab:sota}. (b) Boxplots for driving speed while collecting six selected sequences from the \textit{Boreas} dataset.}
    \label{fig:speed_analysis}
\end{figure}

\noindent
Realistically speaking, it is unlikely that radar sensors are going to completely replace cameras and lidars in perception. As a matter of fact, currently available radars cannot compete with cameras/lidars in terms of object recognition, sampling rates, or Signal to Noise Ratios (SNR). We would expect radars to maintain a strong complementing role in the sensor suite of autonomous platforms. Radars, at least the automotive type, are already widely adopted in the autonomous vehicles market, are still considerably cheaper than lidars, and are much more reliable in bad weather conditions. Some challenges that hinder progress in radar odometry and future recommendations that could alleviate them are discussed here:

\begin{itemize}
    \item{Radars are almost always proposed as a solution for the adverse weather problem and the degradation of lidars and cameras in such scenarios; however, the two most challenging conditions, fog and dust, are the two most underrepresented conditions in the currently available datasets. Research on added synthetic fog has made great progress (e.g., \cite{Hahner2021foggification}) but it would still be preferable to have real data for driving in fog and dust conditions. We, however, recognize the difficulty of anticipating and recording data in such conditions, especially fog.}
    \item{The popular scanning radars from Navtech are known to have a low sampling rate of 4Hz. Given that most publically available datasets were recorded at relatively low driving speeds as shown in Fig. \ref{fig:speed_analysis}}, it means that our radar algorithms are not being put under tests of moderate to high driving speeds. This is problematic as it could mean that our current assessments of the state-of-the-art in radar odometry algorithms are optimistic at best.
    \item{Research in radar perception in general and radar odometry in particular, is missing what the KITTI dataset and leaderboard offers for visual- and lidar-based research; a common reference point for researchers to test and benchmark their work on. Although \textit{Oxford Radar RobotCar} dataset has been filling this role to some extent, \textit{Oxford} dataset is very limited in terms of seasonal diversity and driving speeds. Moreover, it lacks a maintained leaderboard, something that \textit{Boreas} dataset tries to address.} 
    \item{The most popular radar datasets were collected using scanning radars, none of the automotive-based publicly available radar datasets have picked up enough attention to be considered a "radar odometry benchmark", on the contrary, encouraged by its significantly lower cost, a common practice in automotive-radar based research is that for researchers to collect and test on their own unpublished data, which makes it hard to compare and benchmark different odometry approaches that were developed for automotive radars.}
    \item{Finally, the most popular evaluation metrics, the Average Translational Error and Average Rotational Error (see subsection \ref{subsec:driftrate}), were tailored for KITTI dataset; the distances range of (100, 200, 300, 400, 500, 600, 700, and 800) meters doesn't necessarily make sense for other datasets. It could be useful to generalize it to accommodate trajectories with longer or shorter distances.}
\end{itemize}

\section{Conclusion}
\label{sec:conclusion}
\noindent
Radar-based odometry stands as one of the best solutions for estimating the changes in a robot's position and orientation in challenging environments. Radars are well-established in the field of robotics and boast many qualities that make them an important addition to any sensor suite; moreover, radars are getting better, cheaper, and smaller by the day. This article presented a survey of relevant work on the topic of radar odometry, specifically, it focused on radar odometry algorithms intended for autonomous ground vehicles or robots. The current trends in radar odometry research were investigated, including methods, types of sensors, and the utilization of sensor fusion and machine learning techniques. The article also included an overview of the working principles of radar sensors, the standard evaluation metrics in radar odometry, and the publicly available datasets that can be used for radar odometry research. Additionally, a systematic categorization of radar odometry methods found in the literature was presented. Although radar-based state estimation techniques are not new, recent advancements in radar sensor technologies and the increased expectations from autonomy, safety, and all-weather functionalities leave a huge space for more work and development to be done in this field. 

\IEEEtriggeratref{93}
\bibliographystyle{IEEEtran}
\bibliography{myref}
\end{document}